\documentclass[journal,draftcls,onecolumn,12pt,twoside]{IEEEtran}

\usepackage{graphicx} 
\usepackage{booktabs}
\usepackage{amsmath}
\usepackage{mathrsfs}
\usepackage{amsfonts}
\usepackage{amssymb}
\usepackage{epstopdf}
\usepackage{dblfloatfix}
\usepackage[caption=false,font=footnotesize]{subfig}
\usepackage{color}
\usepackage{enumitem}
\usepackage{amsthm}
\usepackage{algorithm,algorithmic}
\usepackage{multirow} 
\usepackage{array}
\usepackage{cite}

\begin{document}

\title{Minimax-optimal decoding of movement goals from local field potentials using complex spectral features}


\author{Marko~Angjelichinoski$^{\ast}$, Taposh Banerjee$^{\dagger}$, John Choi$^{\ddagger}$, \\ Bijan Pesaran$^{\ddagger}$ and Vahid Tarokh$^{\ast}$ 
\thanks{$^{\ast}$ Dptm. of Electrical and Computer Engineering, Duke University (e-mail: $\left\{\mbox{marko.angjelichinoski,vahid.tarokh}\right\}$@duke.edu)}
\thanks{$^{\dagger}$ Dptm. of Electrical and Computer Engineering, University of Texas at San Antonio (e-mail: $\left\{\mbox{taposh.banerjee}\right\}$@utsa.edu)}
\thanks{$^{\ddagger}$ Center for Neural Science, New York University (e-mail: $\left\{\mbox{jc4007, bijan}\right\}$@nyu.edu)}
\thanks{This work was supported by the Army Research Office MURI Contract Number W911NF-16-1-0368.}}


\maketitle

\begin{abstract}
We consider the problem of predicting eye movement goals from local field potentials (LFP) recorded through a multielectrode array in the macaque prefrontal cortex.
The monkey is tasked with performing memory-guided saccades to one of eight targets during which LFP activity is recorded and used to train a decoder.
Previous reports have mainly relied on the spectral amplitude of the LFPs as a feature in the decoding step to limited success, while neglecting the phase without proper theoretical justification.
This paper formulates the problem of decoding eye movement intentions in a statistically optimal framework and uses Gaussian sequence modeling and Pinsker's theorem to generate minimax-optimal estimates of the LFP signals which are later used as features in the decoding step.
The approach is shown to act as a low-pass filter and each LFP in the feature space is represented via its complex Fourier coefficients after appropriate shrinking such that higher frequency components are attenuated; this way, the phase information inherently present in the LFP signal is naturally embedded into the feature space.
The proposed complex spectrum-based decoder achieves prediction accuracy of up to $94\%$ at superficial electrode depths near the surface of the prefrontal cortex, which marks a significant performance improvement over conventional power spectrum-based decoders.
\end{abstract}


\IEEEpeerreviewmaketitle

\section{Introduction}
\label{sec:intro}

Brain-machine interfaces (BMIs) use chronically implanted multielectrode arrays to collect signals from the brain and decode intended motor intentions with the goal of restoring lost motor function \cite{rao_2013,Bensmaia:2014}.
A major challenge in developing effective BMIs is the design of decoders that can reliably distinguish between various motor actions that the subject intends to perform.
In the traditional design of neural decoders, the information-carrying features have been neural spike recordings \cite{Gilja:2012ts,Orsborn:2014ce,Carmena:2003bx}.
However, the ability to detect and collect spikes by chronically embedded microelectrode arrays diminishes over time due to sensor degradation \cite{Stavisky:2015fo}.
Therefore, in recent literature, local field potential (LFP) activity signals are increasingly being considered as a robust and reliable alternative in absence of spike recordings \cite{Linden:2011ck,Pesaran:2018,Einevoll:2013}; the LFP signal is extracted by low-pass filtering ($\leq 1$ kHz) the same wide-band neural signal from which spike recordings are obtained via high-pass filtering and, 
as a result, offers greater long-term decoding reliability, in contrast with spikes. 
Frequency-domain decoding techniques that use the spectrum of the LFP signals as features in the decoding step are a popular choice in the motor decoding literature (see \cite{Jackson:fp} for review).
For instance, the work presented in \cite{Markowitz18412} uses the power spectrum of the LFP to generate the feature space in the detection step.
Although the approach has proven to be effective, albeit to a limited extent, it provides no formal justification for the amplitude information of the Fourier spectrum being sufficient for representing the LFP in feature space.
In other words, power spectrum-based decoders only partially exploit the available decoding information as they neglect the phase information which is inherently present in the LFP signals.

In this paper we cast the problem of decoding neural LFP activity into a statistically optimal framework and use non-parametric regression tools to generate the LFP feature space for decoding; the approach has been initially proposed in \cite{Banerjee1}.
We apply an asymptotically minimax-optimal estimation approach for LFP waveforms based on Gaussian sequence modeling and Pinsker's theorem \cite{Johnstone2012GaussianE}.
The underlying reasoning is that the estimation of the decision surfaces in the feature space using minimax-optimal estimation produces a consistent classifier with worst-case misclassification probability converging to $0$ as the size of the training data set increases \emph{and} as long as the different class-conditional representations of intended motor actions in the feature space remain well separated.
In other words, using minimax estimates as features is firmly justified by the asymptotic performance of the corresponding decoder. 
The approach acts as a low-pass filter followed by shrinkage and thresholding to generate the feature space. Thus, in the approach considered in this paper, the low-band complex spectrum is used which naturally leads to inclusion of the phase information in the decoding process.
We use the data from the experimental setup reported in \cite{Markowitz18412} where two macaque monkeys perform memory-guided saccades to one of eight targets and test the performance of the proposed technique on the problem of decoding intended eye movements directions.
As expected, introducing LFP phase information leads to substantial performance improvements and highly reliable eye movement decoding that significantly outperforms decoders based on power alone, achieving a $94\%$ classification rate.

We also investigate the performance of the complex spectrum-based decoder with increasing depth of the electrodes and across the movement directions and we analyze the impact of the dynamics of decision making process during the memory period when the LFP activity is being recorded. Our findings show that decoding intended motor actions by using the complex low-band representation of the LPFs can have an important implication on the practical design and implementation of real-time, latency-constrained BMIs.  

\section{Methods}

\subsection{Description of the Experiment: Memory-guided Saccades}
\label{sec:experiment}

\begin{figure}
\centering
\includegraphics[scale=0.32]{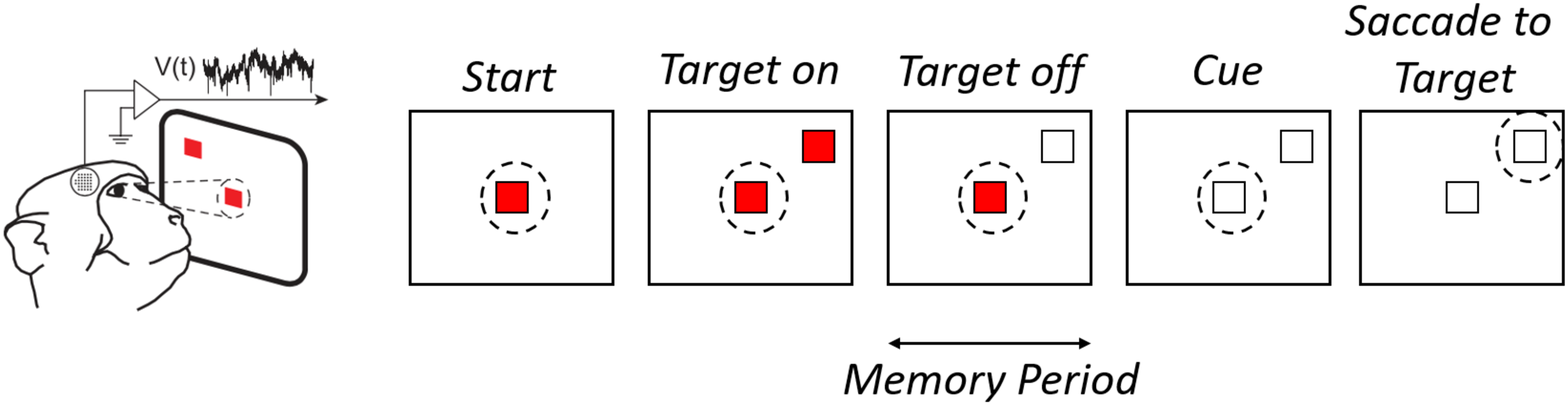}
\caption{Memory-guided saccade experiment. After the subject fixates its gaze on the central fixation target, a randomly selected peripheral target is illuminated for a duration of 0.3 s. After turning the fixation target off, the subject's gaze saccades to the previously illuminated peripheral target location.}
\label{experiment}
\end{figure}

We begin with a brief overview of the experimental setup; we advise the reader to refer to \cite{Markowitz18412} for more details regarding the technical aspects. All experimental procedures were approved by the NYU University Animal Welfare Committee (UAWC).

Two adult male macaque monkeys (\emph{M. mulatta}), referred to herein as Monkey A and Monkey S, were trained to perform a memory-guided saccade task. The setup, illustrated in Fig.~\ref{experiment}, consists of an LCD monitor on which targets are presented to the animal. The monkey initiates a trial by fixating a central target. Once the monkey maintains fixation for a baseline period, one of 8 possible peripheral targets is flashed on the screen for 300 ms. For a random-duration memory period ($1$ - $1.5$ s), the monkey must maintain fixation on the central target until it is turned off. This event cues the monkey to make a saccade to the remembered location of the peripheral target. If the monkey holds his gaze within a window of the true target location, the trial is completed successfully and the monkey receives a reward. Regardless of success or failure, the true target location is reilluminated after the trial as feedback. Targets are drawn from the corners and edge midpoints of a square centered on the fixation target.

An electrode array consisting of $32$ individually movable electrodes (20 mm travel, Gray Matter Research) was implanted into prefrontal cortex (PFC) and used to collect the LFP while the monkeys performed the above task. Signals were recorded at $30$ kHz and downsampled to $1$ kHz. 
A single electrode array topology is henceforth referred to as \emph{electrode depth configuration (EDC)}. 

In our analyses, only LFP segments during the memory periods of a successful trials were used in the decoding experiments. This epoch of the trial is especially interesting since it reflects information storage and generation of the resulting motor response. 


\subsection{Nonparametric Regression Framework for LFP Signal Classification}
\label{sec:background}

In this section we give a brief overview of the nonparametric regression framework which we use for LFP signal estimation and decoding; the framework has been initially considered in \cite{Banerjee1}.
Note that we do not aim for a complete coverage of the underlying theory and we advise the interested reader to refer to \cite{Johnstone2012GaussianE,Tsybakov:2008:INE:1522486} for detailed treatment.

\subsubsection{Outline}

As already mentioned earlier in Section~\ref{sec:intro}, previous efforts have relied on the power spectrum of the LFP signals for feature extraction and decoding of intended motor actions while ignoring the phase information from the spectral representation \cite{Markowitz18412}.
The approach has been shown to be effective (albeit with limited success), proving that the amplitude information indeed stores significant information pertinent to the decision making process.
Nevertheless, the approach still lacks formal justification as the power spectrum yields insufficient representation of the time domain LFP activity. 
Indeed, using only the power spectrum, a time domain signal cannot be fully reconstructed.
The core contribution of this paper is the formulation of the intended eye movement decoding problem within a robust estimation and classification framework grounded on statistical optimality arguments, provided that the LFP waveforms satisfy some mild assumptions.
Building upon the theory of Gaussian sequences and Pinsker's theorem \cite{Johnstone2012GaussianE}, the main result suggests that the statistically optimal way to represent LFP activity and thus generate the feature space where intended motor actions are decoded, should be via the low-band complex frequency spectrum of the LFP data which naturally incorporates the phase information. 


\begin{figure}
\centering
\includegraphics[scale=0.34]{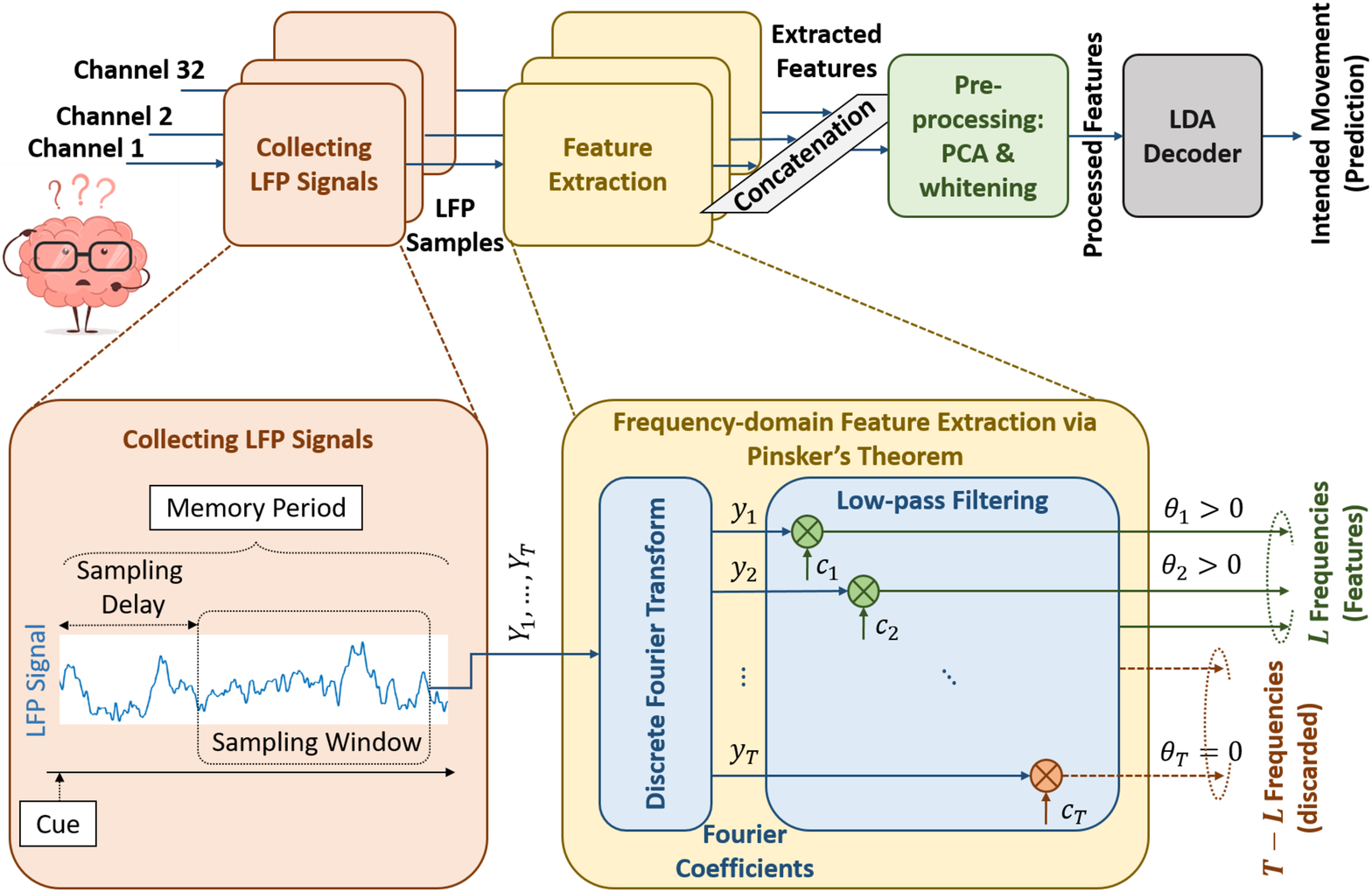}
\caption{Proposed decoder. The top diagram shows all functional blocks relevant for the implementation of the decoder. After collecting a block of LFP samples and extracting the features from them, the feature vectors from each electrode are concatenated and subjected to optional dimensionality reduction and standardization procedure before being fed to an Linear Discriminant Analysis (LDA) decoder. The bottom diagram illustrates the data acquisition and feature extraction procedures (the procedures are applied on a per electrode basis). The sampled time domain LFP signal $Y=(Y_1,...,Y_T)$ is converted into the frequency domain using Discrete Fourier Transform (DFT) and represented via complex Fourier coefficients $y=(y_1,...,y_T)$. The LFP signal is then low-pass filtered in the frequency domain using diagonal filter $C=\text{diag}(c_1,c_2,...,c_T)$ with a cut-off frequency $L\leq T$, i.e., $0<c_l\leq 1$ for $l\leq L$ and $c_l=0$ for $l>L$. According to Pinsker's theorem, the coefficients of the filter should additionally satisfy $c_l\geq c_{l+1}$ for $l=1,...,L$; in other words, Pinsker's frequency domain filter gradually shrinks higher frequencies attenuating their contribution in the representation of the LFP signal. After shrinking, the lowest $L$ non-zero frequency components $\theta = (c_1y_1,c_2y_2,...,c_Ly_L)$ of the spectrum represent the extracted features which are used in the decoding stage.}
\label{LFPfeatureextractionPinsker}
\end{figure}

The proposed feature extraction approach is schematically summarized in Fig.~\ref{LFPfeatureextractionPinsker}; the following subsections provide  details on the different procedures illustrated in Fig.~\ref{LFPfeatureextractionPinsker}.



\subsubsection{Assumptions}
\label{sec:back_assump}

Let $Y_t,t=1,...,T$ represent the discrete time-domain LFP waveform collected from an arbitrary electrode, at an arbitrary depth, during the memory period of a successful trial.
We propose to model the time-domain signal in a {nonparametric regression framework} \cite{Johnstone2012GaussianE,Tsybakov:2008:INE:1522486}:
\begin{equation}\label{eq:LFP_signal}
Y_t = f_{t} + \sigma Z_t,\quad t=1,...,T.
\end{equation}
Here $f$ is a smooth mapping from $[0,1]$ to $\mathbb{R}$, representing the unknown signal, $f_t = f(t/T)$ which is pertinent to the decoding, $Z_t\sim\mathsf{N}(0,1)$ is white Gaussian noise and $\sigma$ captures the standard deviation of the noise component in the observed data.
In a nonparametric setup, $f$ is assumed to belong to a class of smooth functions, denoted by $\mathcal{F}$; in other words, no specific assumptions about the explicit parametric form of $f$ are made.

\begin{figure*}[t]
\centering
\subfloat[Well-separated LFP signal sub-classes]{\includegraphics[scale=0.35]{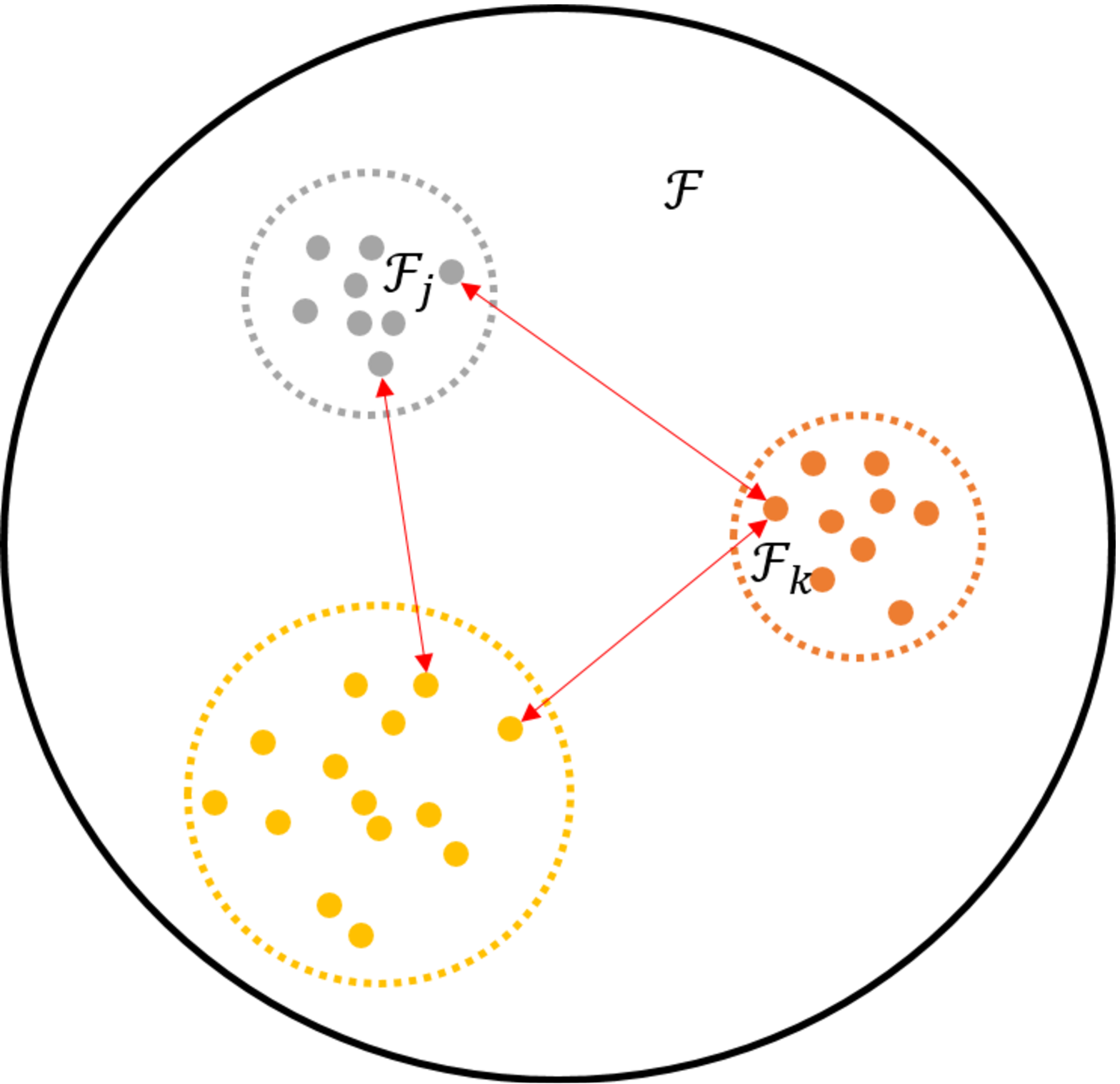}\label{sobolev_space_a}}
\hfil
\subfloat[Overlapping LFP signal sub-classes]{\includegraphics[scale=0.35]{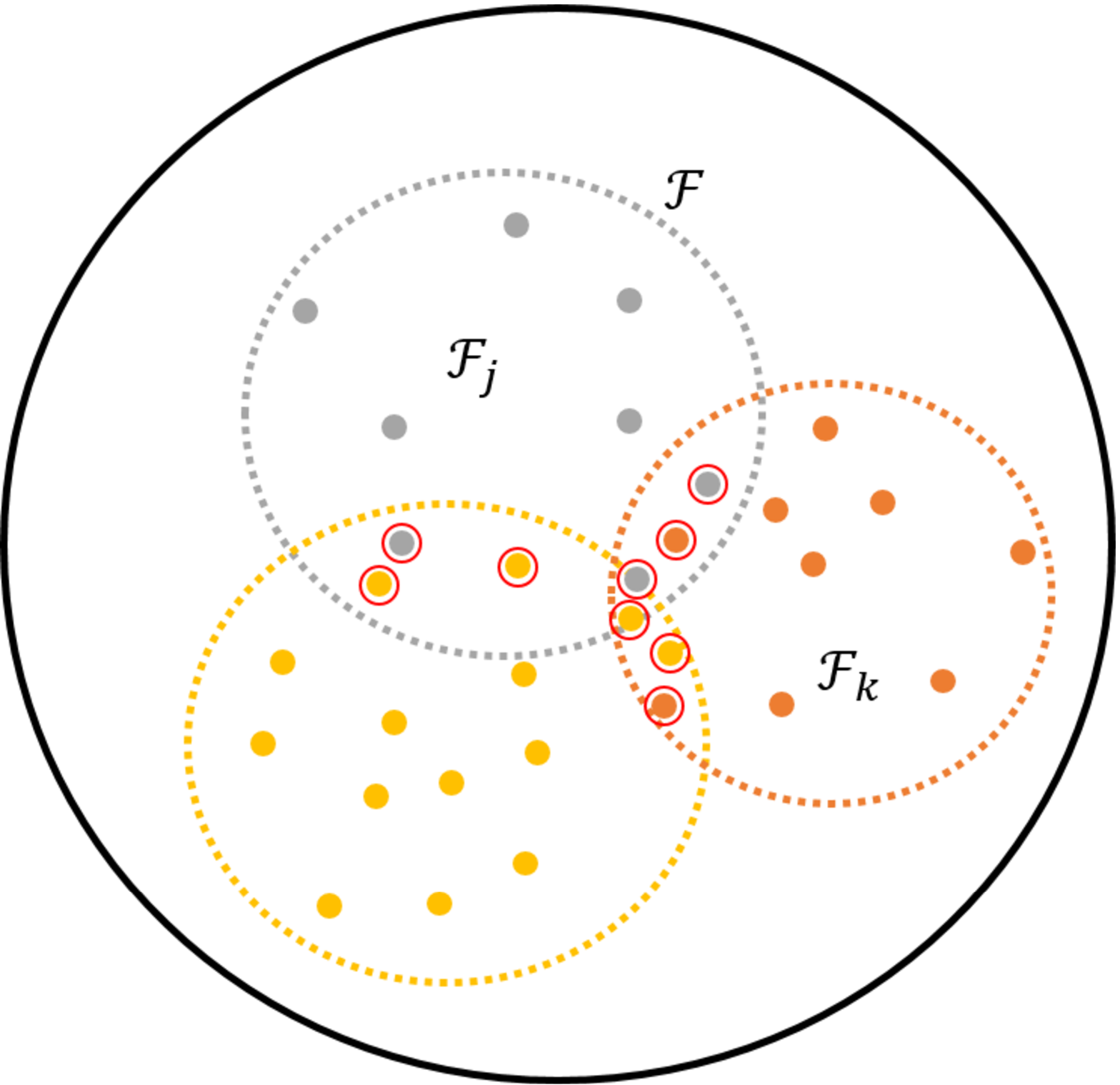}\label{sobolove_space_b}}
\caption{Two-dimensional illustration of LFP signal space configurations. Each LFP signal can be represented via unique function $f$ (filled circles) in a general function class $\mathcal{F}$. The LFP signals corresponding to specific intended action (such as eye movement direction) form a sub-class within the general function class. When the classes are well separated as in (a), the smallest pairwise distances between LFP signal functions in any two sub-classes is non-zero (red arrows) and a carefully tailored decoder (e.g. minimum-distance decoder) can reliably assign given LFP signal function to its corresponding sub-class such that the intended action goals can be reliably decoded. When sub-classes overlap as in (b), some LFP signal functions (in red circles) cannot be unambiguously assigned to their corresponding sub-class; in such case, there is a fundamental upper bound on the performance of any type of decoder, determined by the amount of overlap between sub-classes.}
\label{sobolev_space}
\end{figure*}

We assume that for each target location $k=1,...,K$, with $K=8$ in our case, the function $f$ is \emph{different}.
Moreover, for a given target location, the function $f$ is also expected to vary across different trials such that each target class $k$ forms a sub-class of functions $\mathcal{F}_k$ within the overall function space $\mathcal{F}$, as illustrated in Fig.~\ref{sobolev_space_a}.
This assumption aims to capture the natural variability of the recording conditions as the experiment proceeds; we use the term recording condition to encompass the various different factors that jointly determine the relationship between the LFP activity and the intended motor action such as eye movement direction.
Variations in the recording conditions might be due to, for instance, changes in the electrical properties of the measurement equipment including the electrode array, small drifts of the positions and depths of the electrodes, i.e., changes of the array topology as the experiment proceeds, psycho-physiological changes of the subject, etc.
Therefore, it is reasonable to assume that each individual trial, conducted under the same experimental conditions will produce a {different} function $f$.

Given the above setup, our objective is to find a decoder that maps an arbitrary LFP data sequence $Y_t,t=1,...,T$ into one of the $K=8$ possible target classes, i.e., sub-classes of functions $\mathcal{F}_k,k=1,...,K$.
For reliable detection to be possible at all, the induced sub-classes of functions $\mathcal{F}_k,k=1,...,K$ should be \emph{well separated} in the function space as in Fig.~\ref{sobolev_space_a}.
Formally, we assume that $\mathcal{F}_k\cap\mathcal{F}_j=\emptyset$ for $k\neq j$; this implies that the sub-classes can be easily identified and separated via decision surfaces. 
Provided that this assumption holds, it can be shown \cite{Banerjee1} that if the estimator of the functions $f$ from the available LFP data $Y_t,t=1,...,T$ is such that its worst-case mean squared error (MSE) over the class $\mathcal{F}$ goes to $0$ as the number of successful recordings grows, then a decoder that uses this estimator to determine the decision surfaces is also asymptotically optimal in the sense that its worst case misclassification error converges to $0$; see \cite{Banerjee1} for more details.
This result motivates the use of asymptotically \emph{minimax-optimal} estimator for the functions $f$.
The theory of Gaussian sequences \cite{Johnstone2012GaussianE}, described next, gives a finite-dimensional representation of the minimax-optimal estimators of $f$ and is the theoretical link between the time-domain model \eqref{eq:LFP_signal} and its low-frequency spectral representation which is later on used in the decoding step.

Fig.~\ref{sobolove_space_b}, on the other hand, depicts a situation in which the sub-classes representing different motor actions exhibit significant mutual overlap and the above assumption is not supported; in such case devising an asymptotically optimal decoder with vanishing misclassification rate might be impossible.
An overlap between the induced sub-classes can occur due to variety of reasons, including reduced signal-to-noise ratio, increased electrode depth, clustering successful recordings collected at different EDCs etc.
Some of these aspects are investigated later on in Section~\ref{sec:eval} where we attribute the limited performance of the decoder to a situation reminiscent to Fig.~\ref{sobolove_space_b}.


\subsubsection{Estimation of LFP Signals via Gaussian Sequences}
\label{sec:back_gaussseq}

Given the LFP data $Y_t,t=1,...,T$, we first generate an estimate of the function $f$ using an approach based on the Gaussian sequence representation of model (\ref{eq:LFP_signal}) and Pinsker's theorem \cite{Johnstone2012GaussianE,Tsybakov:2008:INE:1522486}.
The Gaussian sequences allow for finite-dimensional representation of the function $f$ (which typically is infinite-dimensional object and therefore difficult to infer directly) in a conveniently chosen orthonormal basis.
Pinsker's theorem provides a class of asymptotically minimax-optimal estimators of the corresponding finite-dimensional representation of $f$ provided that the function class $\mathcal{F}$ exhibits certain properties.

Let $\phi_l,l\in\mathbb{N}$, be an arbitrary orthonormal basis.
The Gaussian sequence representation is obtained by expanding the discrete-time regression model (\ref{eq:LFP_signal}) into this basis as \cite{Johnstone2012GaussianE}:
\begin{equation}\label{eq:LFP_signal_fourier}
y_l = \theta_l + \frac{\sigma}{\sqrt{T}} z_l,\quad l\in\mathbb{N},
\end{equation}
where $y_l$, $\theta_l$ and $z_l$ are the inner products of the vectors $(Y_1,...,Y_T)$, $(f_1,...,f_T)$ and $(Z_1,...,Z_T)$ with the $l$-th basis function $\phi_l$, respectively.
For LFP signals which exhibit significant temporal smoothness, the Fourier basis is a convenient choice; hence, the Gaussian sequence representation of the LFP data is obtained simply by computing its Fourier coefficients as depicted in Fig.~\ref{LFPfeatureextractionPinsker}:
\begin{equation}
y_l = \frac{1}{T}\sum_{t=1}^T\phi_l(t/T)Y_t,\quad l\in\mathbb{N},
\end{equation}
with
\begin{align}
\phi_1(x) & = 1,\\
\phi_{2l}(x) & = \sqrt{2}\cos(2\pi l x),\quad l\in\mathbb{N},\\
\phi_{2l+1}(x) & = \sqrt{2}\sin(2\pi l x),\quad l\in\mathbb{N}.
\end{align}
Note that in practice we use the equivalent real representation of the complex Fourier series and instead of computing $T$ complex, we actually compute $2T - 1$ real coefficients (one corresponding to the DC and $2(T-1)$ corresponding to the remaining $T-1$ frequency components).

Pinsker's theorem \cite{Johnstone2012GaussianE,Tsybakov:2008:INE:1522486} states that if the sequence of Fourier coefficients $\theta_l,l=1\in\mathbb{N}$, i.e., the vector $\theta = (\theta_1,\theta_2,...)$ lives in an ellipsoid, or, equivalently, $\mathcal{F}$ is a Sobolev class of functions, an asymptotically minimax-optimal estimator of $\theta$ is the linear, diagonal estimator of the form
\begin{equation}\label{eq:diagonal_estimator}
\hat{\theta} = C y
\end{equation}
with $C=\text{diag}(c_1,c_2,...)$ and $y=(y_1,y_2,...)$.
The sequence $c_l,l\in\mathbb{N}$ is non-increasing, gradually decaying to $0$ as $l\rightarrow\infty$; the coefficients $c_l$ are therefore known as \emph{shrinkage coefficients} since they attenuate the impact of higher frequency components on the estimate.
Pinsker's theorem proceeds to give specific analytical form for the shrinkage coefficients which depends on the parameters of the ellipsoid (equivalently, the parameters of the Sobolev class).
Specifically, let $S(\alpha,C)$ be a Sobolev class of functions $f$ defined on $[0,1]$ satisfying $\int_0^1[f^{(\alpha)}(t)]^2 dt\leq C^2$ where $f^{(\alpha)}$ is the derivative of order $\alpha$ of $f$; it can be shown \cite{Johnstone2012GaussianE} that a function $f$ is in a Sobolev class $S(\alpha,\pi^{\alpha}C)$ if and only if its Fourier series coefficients $\theta_l,l\in\mathbb{N}$ are in an ellipsoid, i.e., they satisfy $\sum_l a_l^2\theta_l^2\leq C$ with
\begin{align}
    a_1 & = 1, \\
    a_{2l} & = a_{2l+1}=(2l)^{\alpha},\quad l\in\mathbb{N}.
\end{align}
In such a case, the following linear, diagonal estimator of $\theta_l$ is asymptotically minimax-optimal:
\begin{equation}
    \hat{\theta}_l = \left( 1 - \frac{a_l}{\mu} \right)_{+}y_l,\quad l\in\mathbb{N}.
\end{equation}
The parameters $\mu>0$ and $\alpha$ become design parameters and their values need to be carefully chosen via cross-validation in order to determine the best representation of the LFP waveforms in the decoding space and thus the best prediction performance.
This estimator shrinks the observations $y_l$ by an amount $1-\frac{a_l}{\mu}$ if $\frac{a_l}{\mu}<1$,otherwise it sets the observations to $0$.
As a further simplification, we restrict our attention to \emph{truncation estimator} of the form
\begin{eqnarray}\label{eq:truncation}
\hat{\theta} = \textup{diag}(1_{l\leq L})y.
\end{eqnarray}
Here, $1_{l\leq L}$ is a vector where the first $L<T$ entries are $1$ and the remaining $0$; in other words, our finite dimensional representation of the estimate of $f$ is obtained by simply low-pass filtering the original time-domain sequence to obtain the $L$ dominant components of its spectrum. 
The truncation estimator asymptotically also achieves the minimax rate of convergence, see \cite{Johnstone2012GaussianE}, it is asymptotically minimax-optimal and can be viewed as a special case of Pinsker's estimator; it is also simpler to implement than the Pinsker's estimator since it introduces only a single design parameter, namely the cut-off frequency $L$.

\begin{figure}[t]
\includegraphics[scale=0.575]{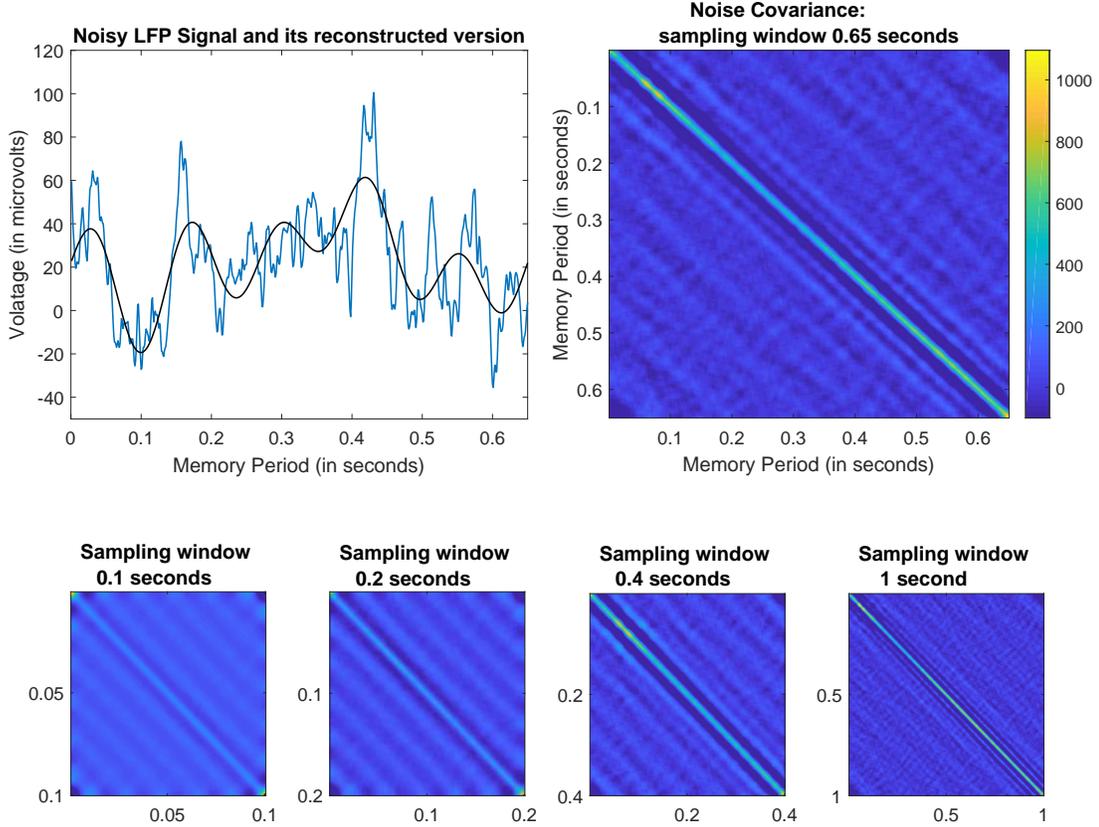}
\caption{LFP signal and noise analysis. An arbitrary electrode at an arbitrary depth was chosen. The top-left plot depicts a noisy LFP signal (blue line) recorded during an arbitrary successful trial and collected during the memory period with sampling window of duration $650$ milliseconds and sampling delay of $D=0$ seconds (tracing starts immediately after the central fixation target goes off, see Fig.~\ref{experiment}) and the robust estimate of the true LFP signal (black line) obtained via inverse Fourier transform using $L=5$ coefficients. For the same electrode, we evaluated the empirical, time-domain covariance matrix of the remaining noise after reconstructing the LFP estimate and subtracting it from the original signal and averaging over all successful trials recorded at the given depth. The top-right plot shows the covariance matrix for sampling window of $650$ milliseconds whereas the bottom plots depict the covariance matrix computed for several sampling window sizes. 
}
\label{LFPwave}
\end{figure}

\subsubsection{LFP Signal Analysis}
One of the key assumptions introduced in the model (\ref{eq:LFP_signal}) is the noise independence; adequate adherence of the LFP data to this assumptions plays a critical role for the realized performance of the decoder.
We therefore investigate the properties of the LFP signal noise.
The top-left plot in Fig.~\ref{LFPwave} depicts LFP data sequence $Y_t,t=1,...,T$ collected during a successful trial at an arbitrary EDC from an arbitrary electrode and the reconstructed time-domain version of the estimated waveform using $L=5$ Fourier coefficients.
The top-right plot depicts the empirical covariance matrix of the remaining noise after subtracting the reconstructed signal (black line in Fig.~\ref{LFPwave}) from the LFP data (blue line in Fig.~\ref{LFPwave}), computed by averaging across all trials collected from the same electrode at the given depth configuration. 
We see that the covariance matrix is dominated by the main diagonal, i.e., the LFP samples are weakly correlated which justifies the noise model in (\ref{eq:LFP_signal}); this is an important finding as the data adherence to the assumptions of Pinsker's theorem plays crucial role for the realized performance of the decoder.
Albeit not depicted, we note that the diagonally-dominated structure of the time-domain covariance matrix remains consistent across electrodes and across depth configurations.
However, the bottom plots which depict the covaraince matrix for various sampling windows sizes, ranging from $0.1$ to $1$ second, show that the weak noise correlation can be reasonably claimed for larger sampling window sizes; in other words, on finer time scales, the independence assumption in (\ref{eq:LFP_signal}) becomes increasingly invalid.

\subsubsection{Linear Discriminant Analysis for LFP Signals}
\label{sec:back_lda}

$\hat{\theta}$ obtained via (\ref{eq:truncation}) is now our feature over which we train the decoder. 
Obviously, doing the decoding in the frequency-domain is more convenient practical choice compared to the time-domain which involves computing the inverse Fourier transform and obtaining the estimated sequence $( \hat{f}_1,...,\hat{f}_T )$.
This is due to the reduced dimensionality of the problem in the frequency-domain since $L<T$; in fact, as we will see later in Section~\ref{sec:eval}, $L\ll T$, i.e., typically, we need only the very few low frequencies for adequate representation of the LFP waveforms.

The estimate $\hat{\theta}$, under the model (\ref{eq:LFP_signal}) is postulated to be a multivariate Gaussian random variable with mean vector $\mu$ and covariance matrix $\Sigma$: 
\begin{equation}\label{eq:theta_gauss}
\hat{\theta}\sim\mathsf{N}(\mu_k,\Sigma).
\end{equation}
The mean vectors are different for each target location $k=1,...,K$ while the covariance matrices are assumed to be equal (shared).
Hence, linear discriminant analysis (LDA) is suitable for decoding.

\subsection{The Decoding Algorithm}
\label{sec:DecodAlg}

A pictorial illustration of the different procedures included in the decoding algorithm is presented in the top diagram in Fig.~\ref{LFPfeatureextractionPinsker}.
Multichannel LFP data as described in Section~\ref{sec:experiment} is collected from the memory period of the task.
The duration of the sampling window, i.e., the number of LFP samples $T$ and its start within the memory period, i.e., the delay $D$ are design parameters.
This is motivated by the intuition that not all samples of the memory period carry equal amount of information relevant to the decision making process. 
We first compute the Fourier coefficients for each channel individually after which they are concatenated into a single, large feature vector.
We then apply Principle Component Analysis (PCA) and Zero Component Analysis (ZCA) to reduce the dimension of the original feature space and standardize the features into a lower-dimensional subspace.
The number of retained principal modes $P$ becomes a free parameter that is optimized in addition to the other free parameters.

We summarize the decoding algorithm:
\begin{enumerate}
    \item Fix the free parameters $T,D,L,P$.
    \item \emph{Data preparation.} Form $32\times T$ data matrix for each trial by placing the $T$ LFP time-domain samples collected from each channel as rows of the matrix.
    \item \emph{Feature extraction via low-pass filtering}. Compute $L$ Fourier coefficients for each row of the data matrix $2L-1$ real coefficients, corresponding to one DC component, $L-1$ sines and $L-1$ cosines.
    \item \emph{Feature space formation}. Append the Fourier coefficients of all $32$ channels into a single high-dimensional feature vector of length $32(2L-1)$; this is the dimension of the original feature space.
    \item \emph{Data pre-processing}. Use PCA and ZCA to project the high-dimensional feature vector onto $P$ principal modes and standardize the features, respectively; in principle, this is an optional procedure.
    \item \emph{Decoding intended movements}. Train the LDA decoder.
\end{enumerate}
The performance of the decoder is optimized over the free parameters via cross-validation.

\subsection{Decoding Across Multiple Electrode Depths}
\label{sec:exp_edc}

A crucial aspect of the experiment concerns the depths of the individual electrodes; these depths are varied across different sessions and more details on how this is done can be found in \cite{Markowitz18412}.
Multiple trials are performed for a given EDC.
As described later on in Section~\ref{sec:eval_data}, the number of trials is approximately uniformly distributed across different EDCs.
For Monkey A in this experiment however, there is a particular EDC at which multiple sessions were conducted and therefore substantially higher number of trials were recorded; due to the abundance of data compared to other EDCs, this data set was used in \cite{Banerjee1}.
All other EDCs however, typically correspond to a single recording session and therefore the number of available trials per EDC is substantially lower, often so small that no meaningful training procedure can be performed.
This prevents us from applying the LDA to each individual EDC as it leads to an ill-posed problem.
To circumvent this issue and to be able to test our decoder across varying electrode depths, we rely on the following reasoning: similar EDCs produce similar function/feature spaces; more formally, given an arbitrary pair of EDCs with similar configuration vectors (describing the depths of the electrodes), we expect that the function/feature spaces will be similar.
Hence, data sets of EDCs which are close to each other in Euclidean sense can be grouped, i.e., clustered together to form one large data set.
The data clustering procedure is summarized as follows:
\begin{enumerate}
    \item fix the concurrent EDC where the decoding performance is being investigated;
    \item define a clustering window as the minimum number of trials per EDC;
    \item append the trials collected at the concurrent EDC;
    \item until the clustering window is full, append trials from neighboring EDCs, i.e.,  EDCs whose depth vectors are the closest in Euclidean distance from the depth vector of the concurrent EDC.
\end{enumerate}
By setting the size of the clustering window to an appropriate value, the clustering procedure yields data clusters of sufficient training sizes.
Note that the procedure of clustering trials collected at different depths will likely violate our previous assumption about non-overlapping function classes and is therefore expected to lead to degradation of the decoding performance due to increasingly overlapping functions sub-classes as we group data patches collected across heterogeneous EDCs; see Fig.~\ref{sobolev_space}.

\section{Results}
\label{sec:eval}

We first describe the collected data from the two subjects Monkey A and Monkey S and then we evaluate the performance of the complex spectrum decoder. 

\subsection{Data Description}
\label{sec:eval_data}

\begin{figure*}[t]
\centering
\subfloat[Monkey A]{\includegraphics[scale=0.5]{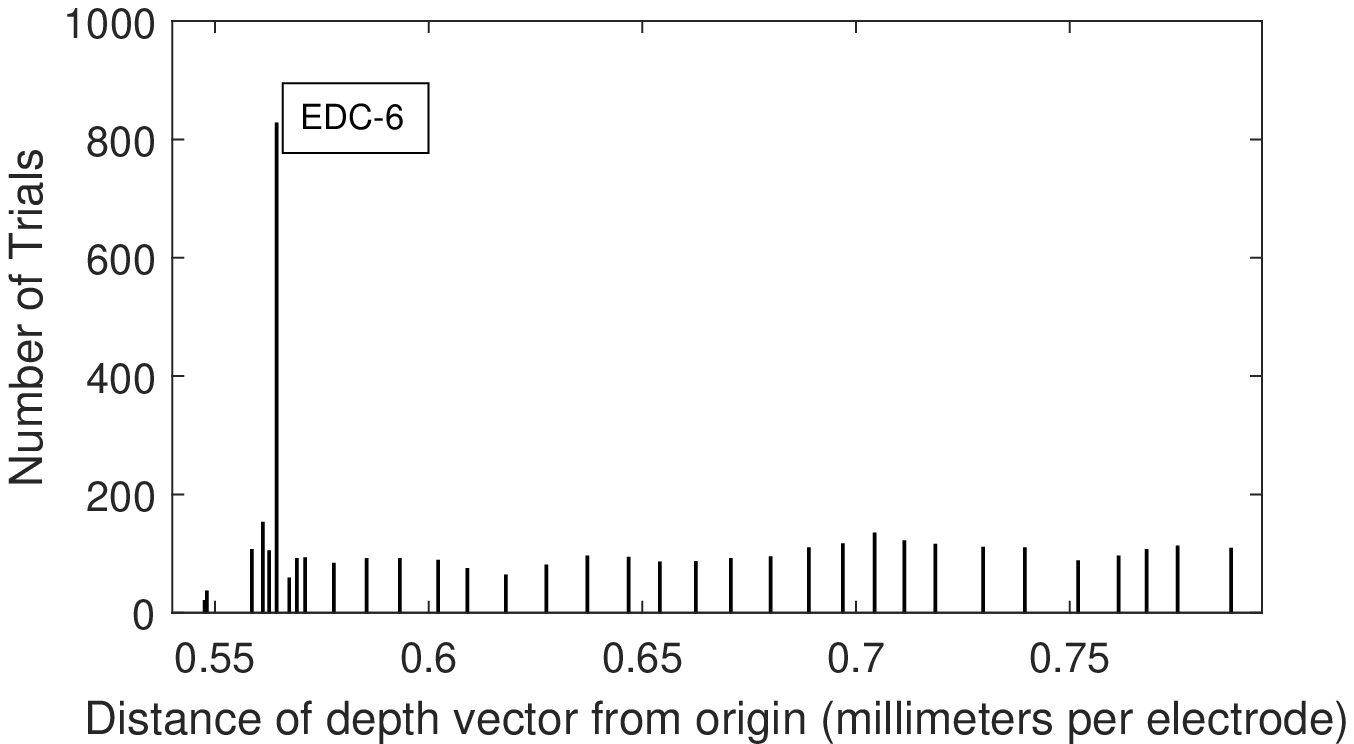}\label{data_a}}
\hfil
\subfloat[Monkey S]{\includegraphics[scale=0.5]{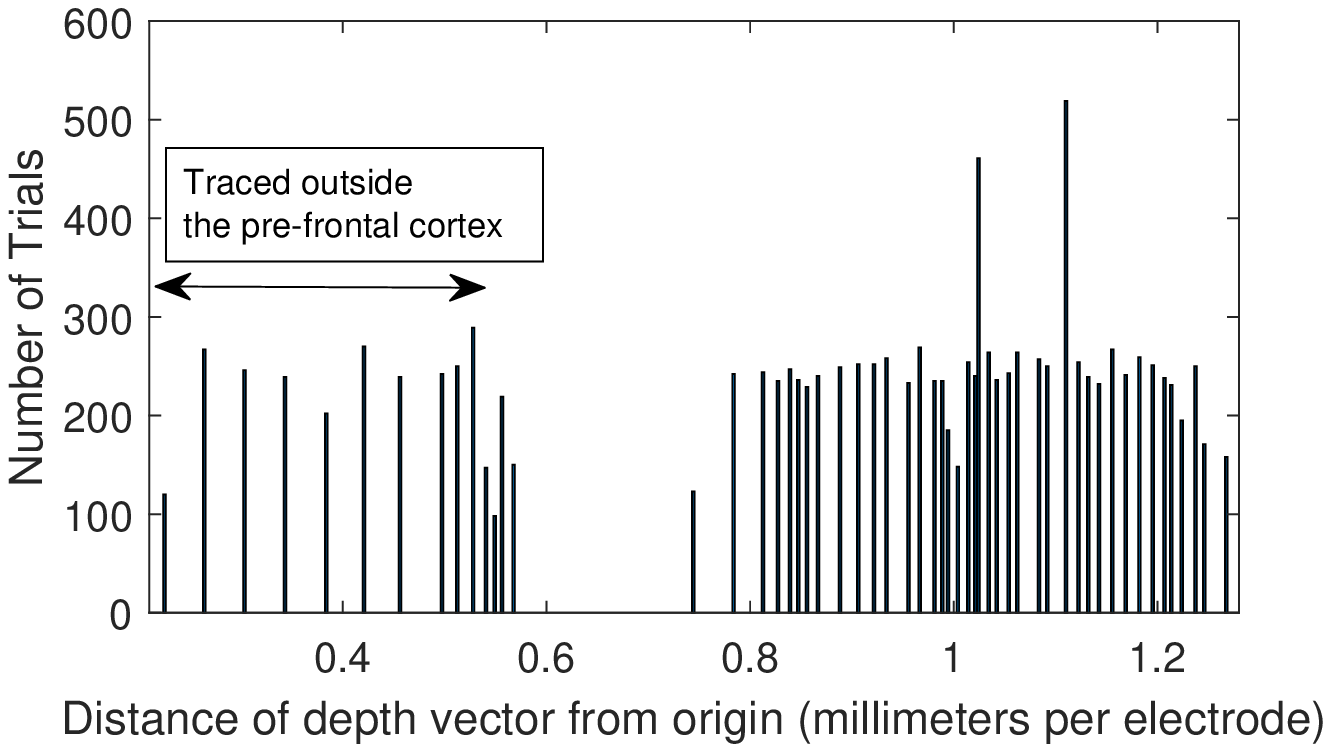}\label{data_b}}
\caption{The distribution of successful trials across EDCs. EDCs are denoted by the ordering with respect to their depths vector from the origin.}
\label{data}
\end{figure*}

Multiple trials are performed for a given EDC; each EDC is described by a $32$-dimensional depth vector comprising the depths of the individual electrodes.
The number of successful trials collected for each EDC is presented in Fig.~\ref{data} with the horizontal axis denoting the Euclidean distance of the EDC vector in millimeters per electrode.  

The LFPs have been collected over $34$ and $55$ unique EDCs for Monkey A and Monkey S, respectively.
From Fig.~\ref{data}, we observe that the trials are approximately uniformly distributed across EDCs with the average number of trials per EDC standing at approximately $94$ and $230$ for Monkey A and Monkey S, respectively. 
There is however a particular EDC in the Monkey A traces, denoted as ``EDC-$6$'' in Fig.~\ref{data_a} with $827$ trials, collected over $10$ recording sessions; due to the abundance of trials compared to other EDCs, this data batch was used in the previous work \cite{Banerjee1}. 
We note that except for EDC-$6$ in Monkey A, the number of trials for all other EDCs is quite small compared to the overall dimension of the features, which even after PCA will still be above $100$, see \cite{Banerjee1} and Section~\ref{sec:eval_res}.
To resolve this issue and be able to inquire the performance of the decoder across different sampling depths, we cluster trails from multiple EDCs based on the Euclidean proximity of the corresponding depth vectors as described in Section II.  

\subsection{Evaluations}
\label{sec:eval_res}


We use leave-one-out cross-validation to optimize the decoder over the free parameters. 
We investigate the impact of electrode depth, eye movement direction and the dynamics of the memory period onto the performance of the decoder.
Note that most of the analyses have been conducted using the values of the free parameters optimized over the best performing data set; this does not necessarily mean the same values are optimal for other data sets and configurations. Nevertheless, the variations in performance are minor.

\subsubsection{Impact of Electrode Depths}

\begin{figure*}[t]
\centering
\subfloat[Monkey A]{\includegraphics[scale=0.5]{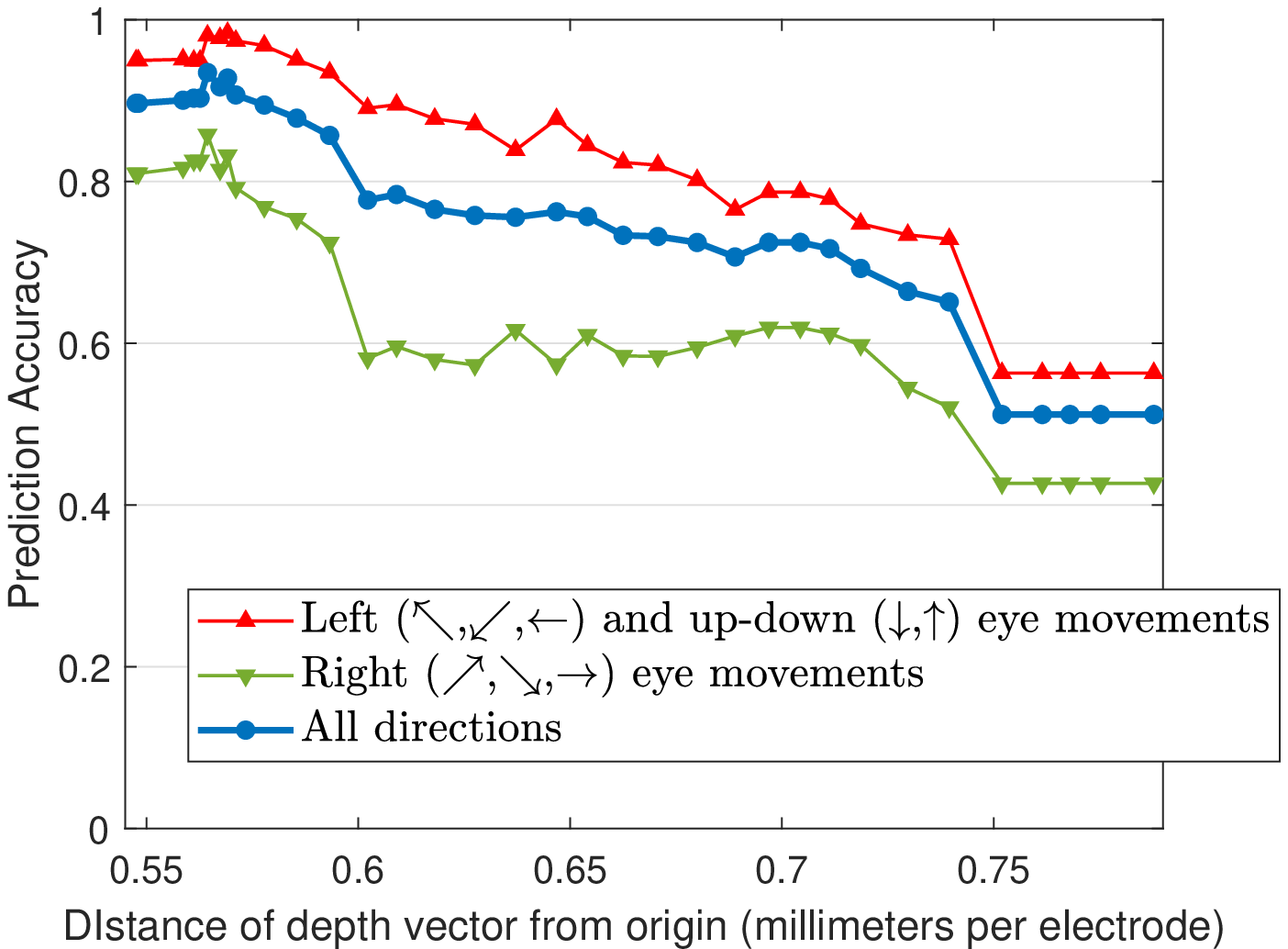}\label{results3a}}
\hfil
\subfloat[Monkey S]{\includegraphics[scale=0.5]{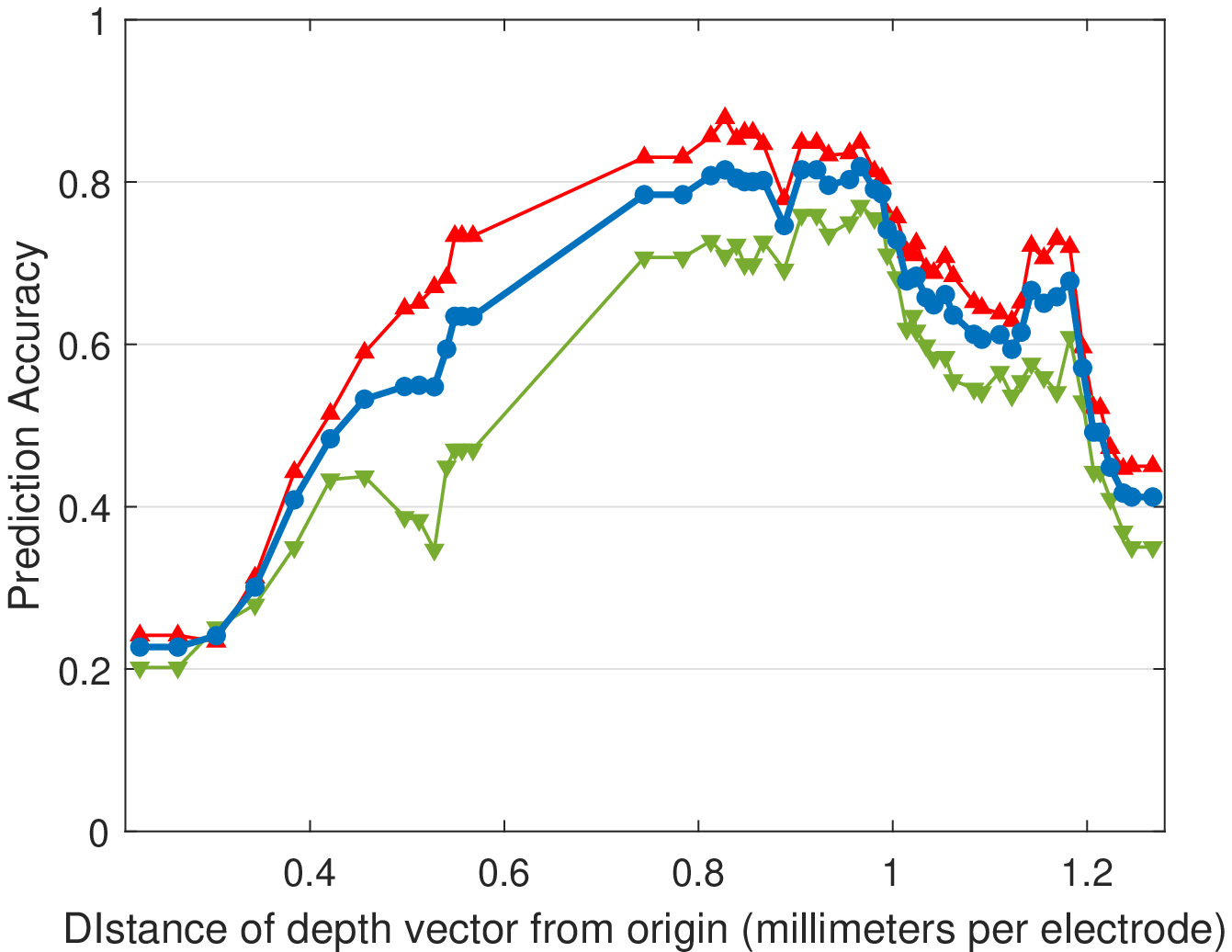}\label{results3b}}
\hfil
\subfloat[Monkey A]{\includegraphics[scale=0.5]{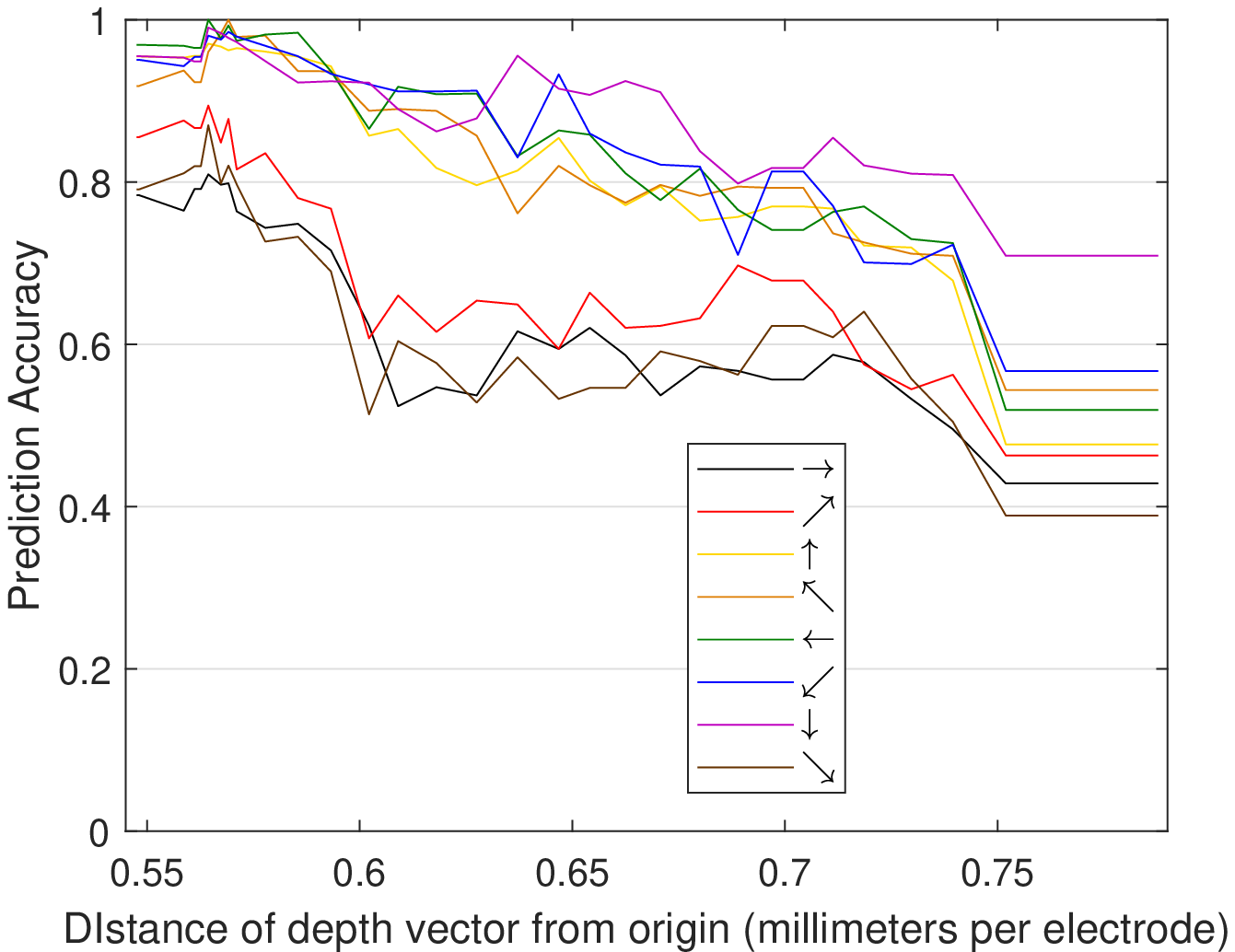}\label{depths_targets_a}}
\hfil
\subfloat[Monkey S]{\includegraphics[scale=0.5]{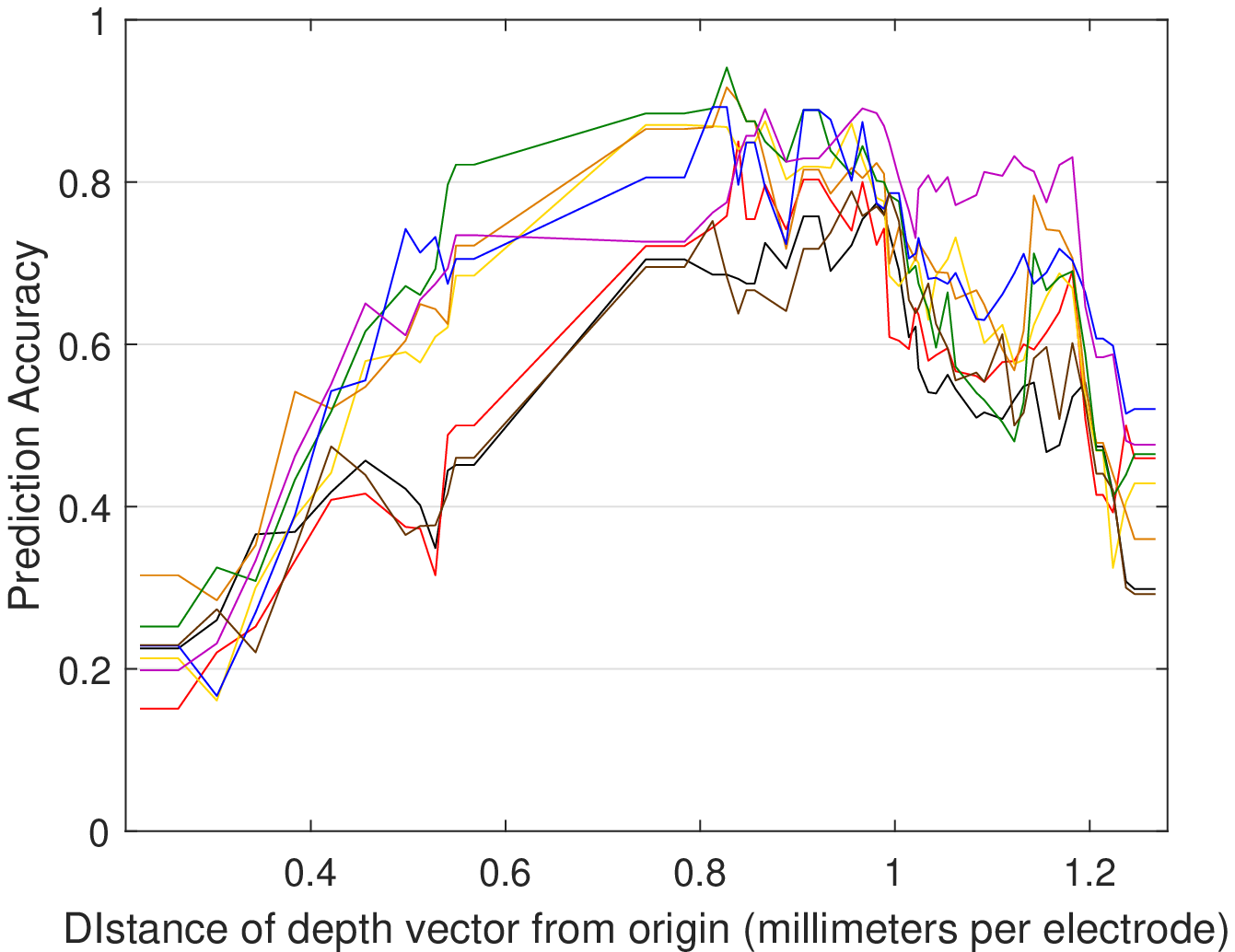}\label{depths_targets_b}}
\caption{Decoding performance across Electrode Depth Configurations (the legend applies to both figures). The optimal number of Fourier coefficients and principal modes were found to be $L=4$ and $P=187$ respectively. The sampling delay within the memory period is set as $D=0$, i.e., we start sampling at the very beginning of the memory period while the optimal sampling window size is set at $T=650$ samples, i.e., $650$ milliseconds. The clustering window size is set to $900$. The free parameters have been optimized over the best performing data cluster. (a),(b) depict average performance across all directions and across the ipsilateral and contralateral (plus up/down) directions. (c),(d) depict performance for each direction individually.}
\label{results_depth}
\end{figure*}

\begin{figure*}[t]
\subfloat[Monkey A]{\includegraphics[scale=0.38]{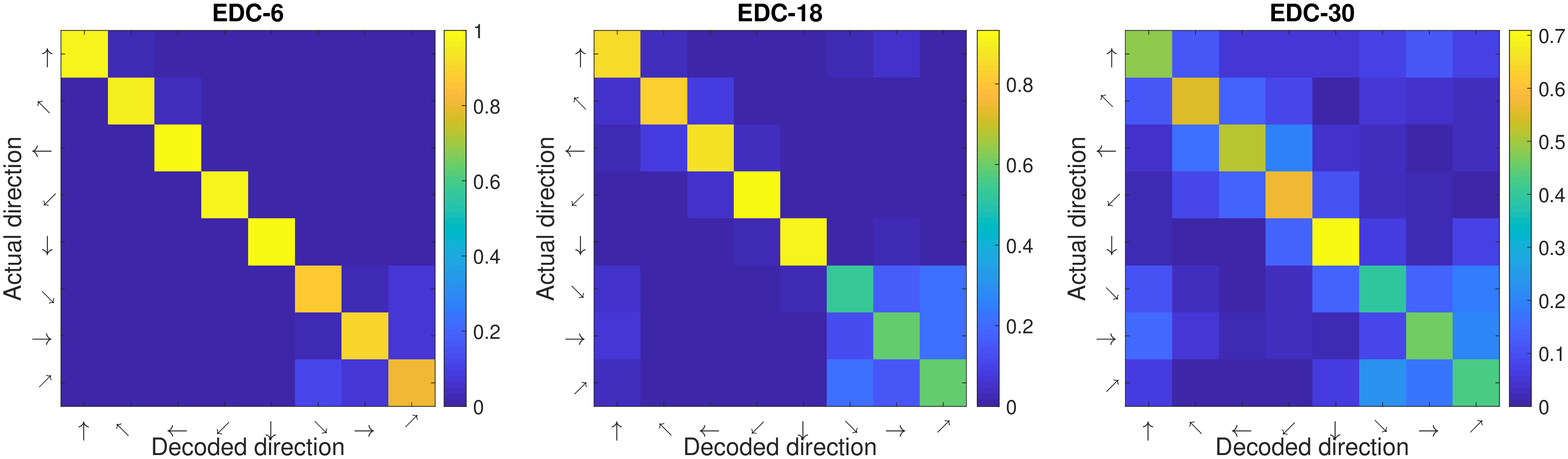}\label{hitmaps_a}}
\hfil
\subfloat[Monkey S]{\includegraphics[scale=0.38]{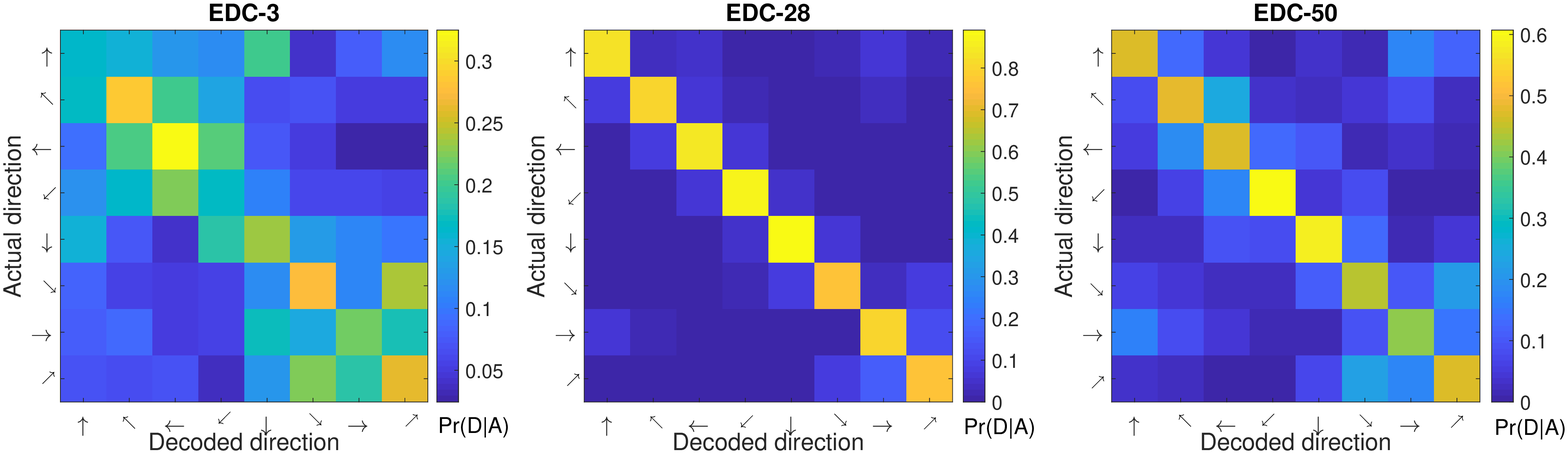}\label{hitmaps_b}}
\caption{Confusion matrices for representative EDCs. The free parameters are optimized over the best performing data set (see caption of Fig.~\ref{results_depth}). Entry $(i,j)$ in each confusion matrix corresponds to an empirical estimate of the conditional probability of decoding direction $j$ provided that the true direction is $i$.}
\label{hitmaps}
\end{figure*}

Fig.~\ref{results_depth} depicts the performance the decoder across electrode depths.
The main conclusion that can be drawn from the analysis is that the decoder achieves the best performance near the surface of the prefrontal cortex.
It should be noted that with Monkey S, recording started while the electrodes were still outside the prefrontal cortex, i.e., before any multiunit activity was detected; this is why the decoder exhibits low performance for the initial EDCs.
We also observe that for Monkey A, the decoder peaks at $94\%$ prediction accuracy which is obtained for EDC-6 in addition to the closest neighboring data patch.
Similar performance (i.e., above $92\%$ up to $94\%$) can be obtained for EDCs in the vicinity of EDC-6 after fine tuning the free parameters.
In Monkey S, the decoder peaks at $82\%$, again for depths near the surface of the prefrontal cortex.
We observe that as we increase the actual depth of the electrodes the performance degrades for both subjects.
Although the trend of performance degradation with increasing electrode depth is an intuitively expected phenomenon and has been also reported in previous analyses \cite{Markowitz18412}, we remain cautious with respect to the extent to which the actual performance can be attributed solely to the increasing electrode depths.
In particular, apart from the decreased signal-to-noise ratio as we approach white matter with increased electrode depths, the performance degradation can in part be associated with an increased overlap between the function sub-classes in the function space as we group multiple small data patches collected across heterogeneous EDCs.
Recall that recording conditions evolve as the experiment progresses, leading to displacement of the function sub-classes across the multidimensional feature space, eventually leading to an overlap between sub-classes associated with different eye movement directions.
In light of this, note that there is a decoding performance gap between Monkey A and Monkey S; while the decoder achieves $94\%$ accuracy for Monkey A, it only achieves $82\%$ for Monkey S.
As also observed in earlier works \cite{Banerjee1,Markowitz18412}, this is due to the homogeneous nature of the EDC-6 data set.
Specifically, as we can see in Fig.~\ref{data_b}, no data set in Monkey S has more than $600$ trials; as a result, the data clusters that were formed to train the decoder are highly heterogeneous, likely leading to an overlap between the function classes corresponding to different eye movement directions as in Fig.\ref{sobolove_space_b}.

\subsubsection{Impact of Eye Movement Direction}


Fig.~\ref{results_depth} also shows the performance range of the decoder across the depths using the average performance computed for contralateral (including up/down) and ipsilateral eye movement directions separately, as in Fig.~\ref{results3a} and \ref{results3b}, as well the complete performance range across each direction separately, see Fig.~\ref{depths_targets_a} and ~\ref{depths_targets_b}. 
We observe that there is a substantial performance gap between the contralateral and the ipsilateral directions - a phenomenon which has been also observed and reported previously \cite{Markowitz18412}.
To investigate further, in Fig.~\ref{hitmaps} we show the confusion maps as $8\times 8$ matrices (where the rows correspond to the true eye movement direction while the rows give the decoded one) for three representative EDCs for both Monkey A and Monkey S.
We clearly observe (especially as we go deeper) that the ipsilateral eye movements are prone to more misclassifications; at superficial depths where the decoder attains its best performance, the contralateral and the up/down directions are decoded almost perfectly and the overall error is dominated by the misclassifications related to the left directions.
In addition, we also observe that the confusion maps tend to have block-diagonal structure with the contralateral (plus up/down) directions forming one block and the ipsilateral targets forming another; hence, majority of errors especially at superficial depths remain within the same block, i.e., right/left eye movement directions are predominantly confused with other right/left directions with the inter-block error rate being negligible.
However, as the depth of the electrodes increases and due to the drift of recording conditions over time, we observe increased confusion between blocks - which is likely a result of the increased overlap between function sub-classes.


\subsubsection{Comparison with Power Spectrum-based Decoder}

\begin{figure*}[t]
\centering
\subfloat[Monkey A]{\includegraphics[scale=0.5]{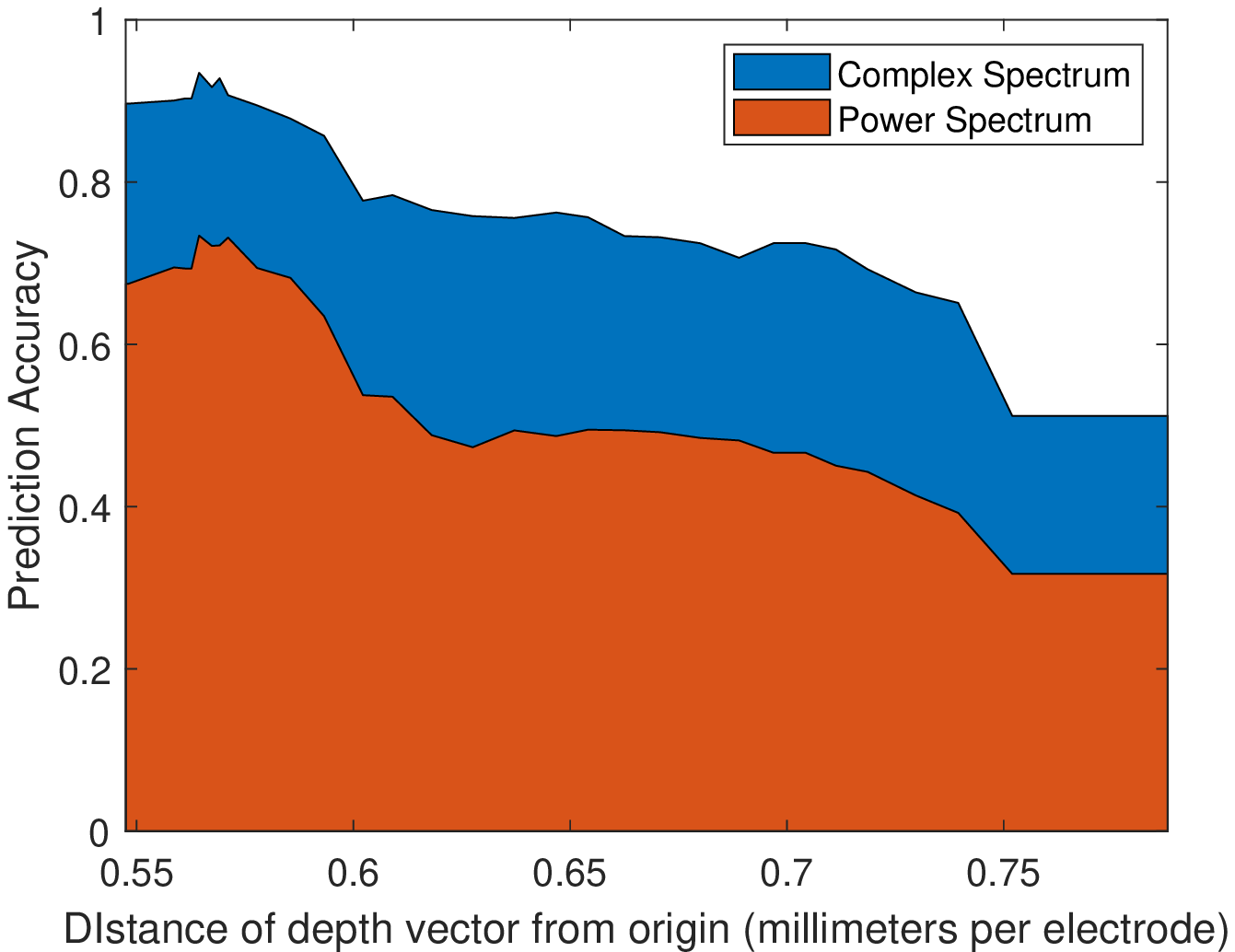}\label{power_depth_a}}
\hfil
\subfloat[Monkey A]{\includegraphics[scale=0.5]{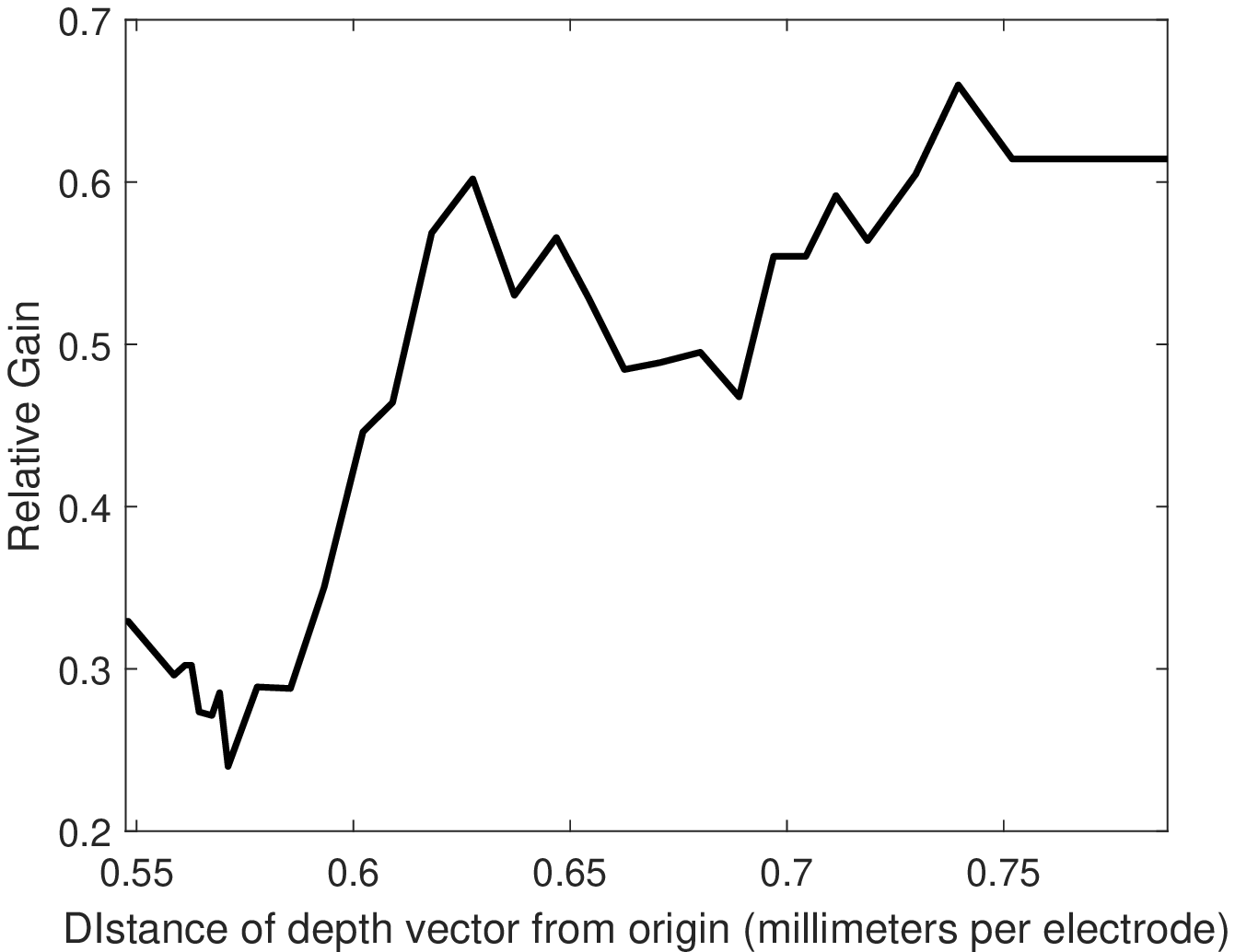}\label{power_depth_b}}
\hfil
\subfloat[Monkey S]{\includegraphics[scale=0.5]{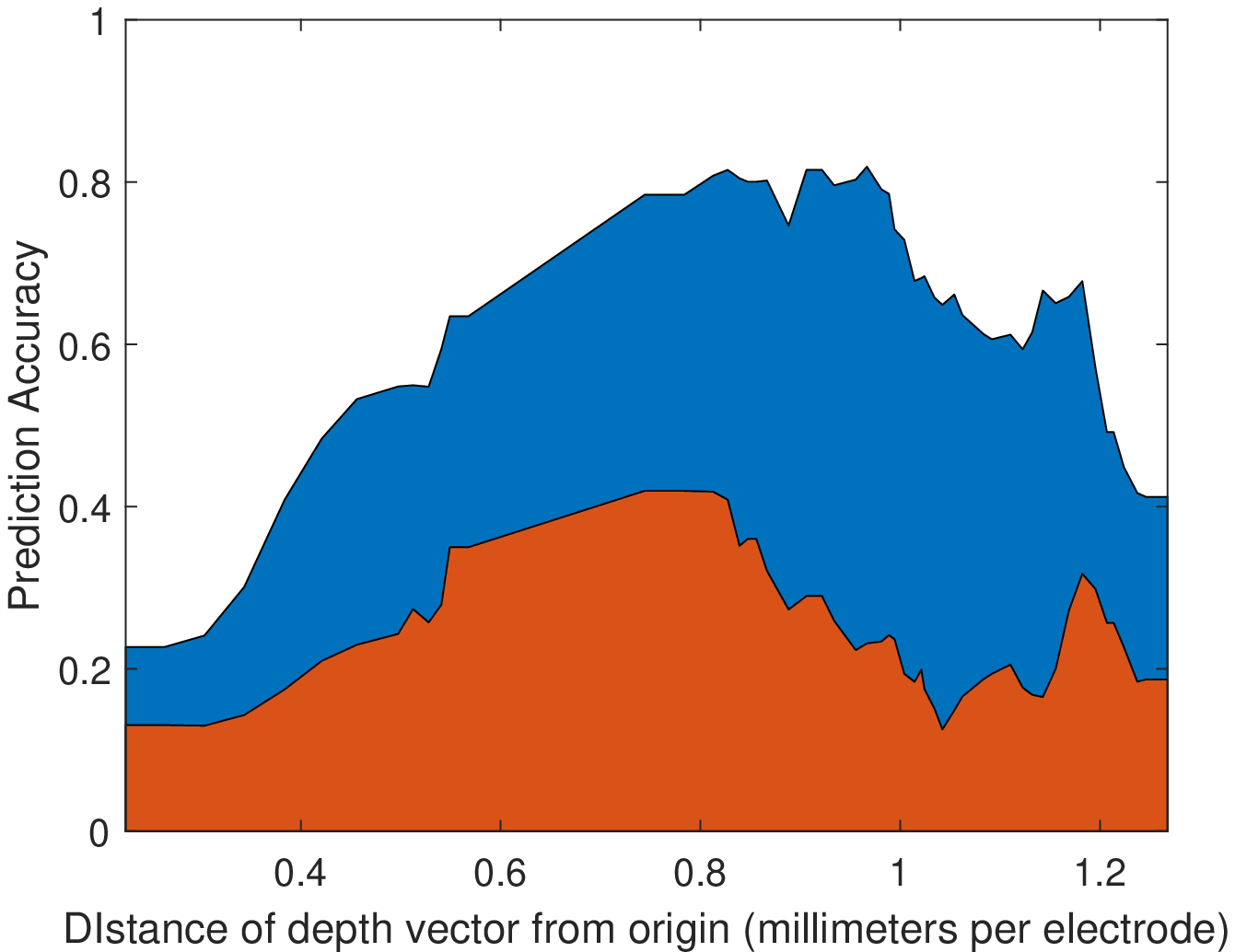}\label{power_depth_c}}
\hfil
\subfloat[Monkey S]{\includegraphics[scale=0.5]{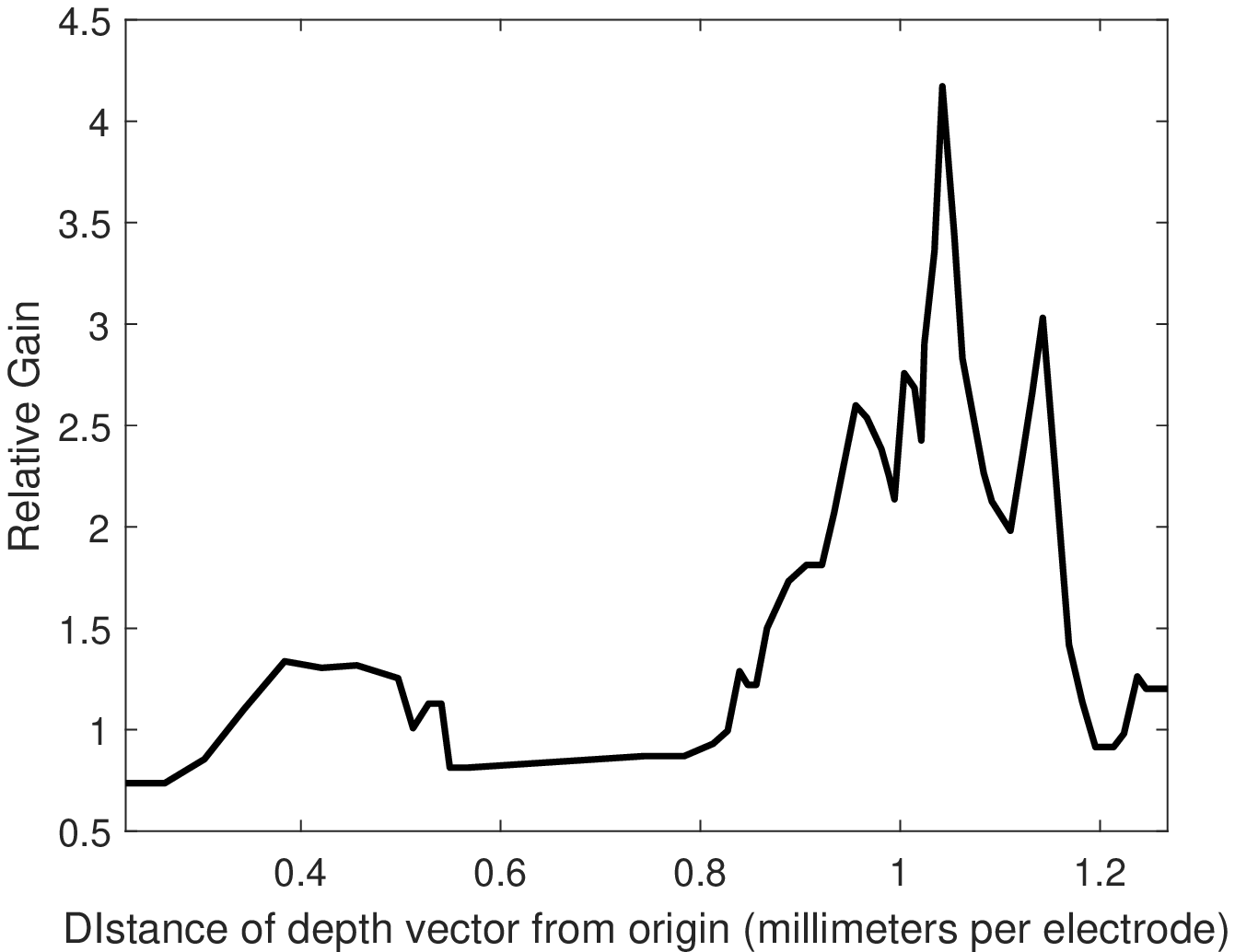}\label{power_depth_d}}
\caption{Comparison between decoding performance of complex and power spectrum based decoders. The legend in (a) also applies to (c). The relative gain in (b) and (d) is computed relative to the performance achieved by the power spectrum decoder. The optimal number of Fourier coefficients and principal modes were found to be $L=4$ and $P=187$ (complex spectrum based decoder) or $P=100$ (power spectrum based decoder), respectively and they are optimized over the best performing data set. All other free parameters remain the same as in caption of Fig.~\ref{results_depth}.}
\label{power_spectrum_depths}
\end{figure*}

The performance of the proposed decoder, especially near the surface of the prefrontal cortex, is attributed to the fact that the decoder uses the complex spectrum and exploits the phase information from the Fourier coefficients as opposed to conventional power spectrum-based techniques.
In Fig.~\ref{power_spectrum_depths} we compare the performance of our complex with a power spectrum decoder; the free parameters remain the same as before, namely the sampling window size is $T=650$ samples (i.e., milliseconds), the number of Fourier coefficients is $L=4$ and the number of retained principal modes is $P=187$ for the complex spectrum decoder and $P=100$ for the power spectrum decoder.
We see that with the same number of Fourier coefficients per electrode, the complex spectrum decoder proposed in this paper consistently outperforms the conventional power spectrum decoder.
Fig.~\ref{power_depth_b} and \ref{power_depth_d} depict the relative gain of the complex over the power spectrum decoder.
We observe that the gain, defined in relative terms with respect to the performance initially achieved by the power spectrum decoder for the same number of Fourier coefficients per channel, tends to increase with increased electrode depths.
This outcome might suggest that the phase information plays an increasingly important role in the success of the decoder when the sub-classes associated with different directions in decoding space overlap.

\subsubsection{Impact of Sampling Window and Sampling Delay}

\begin{figure}[t]
\centering
\includegraphics[scale=0.6]{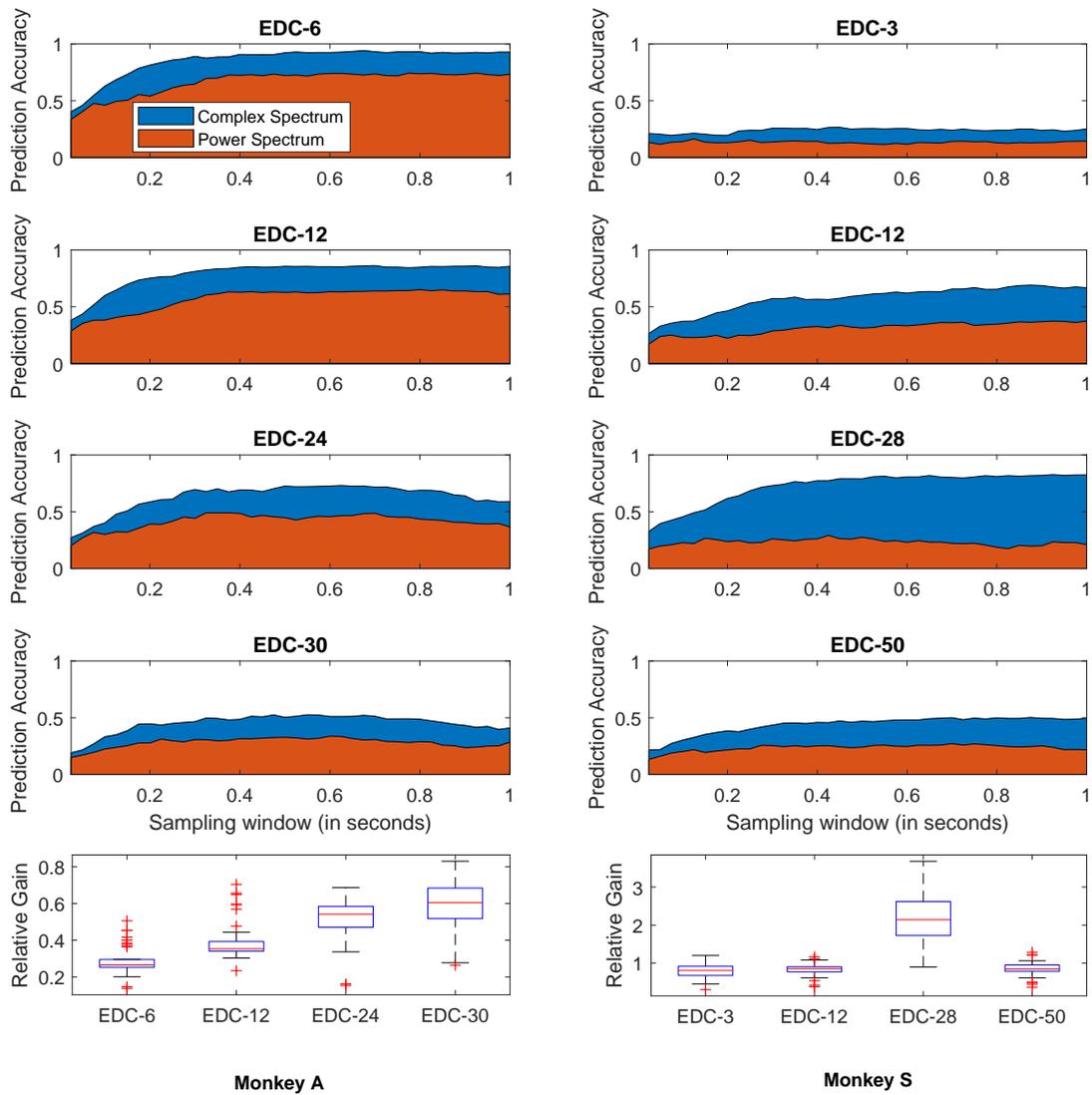}
\caption{Performance across sampling windows for representative EDCs. The sampling delay is set to $0$, i.e., sampling begins immediately after the fixation target goes off. The remaining free parameters remain the same as in caption of Fig.~\ref{power_spectrum_depths}. The box plots depict the distribution of the relative performance gain, computed with respect to the performance achieved by the power spectrum-based decoder, across the sampling window sizes.}
\label{sampling_window}
\end{figure}

\begin{figure}
\centering
\includegraphics[scale=0.6]{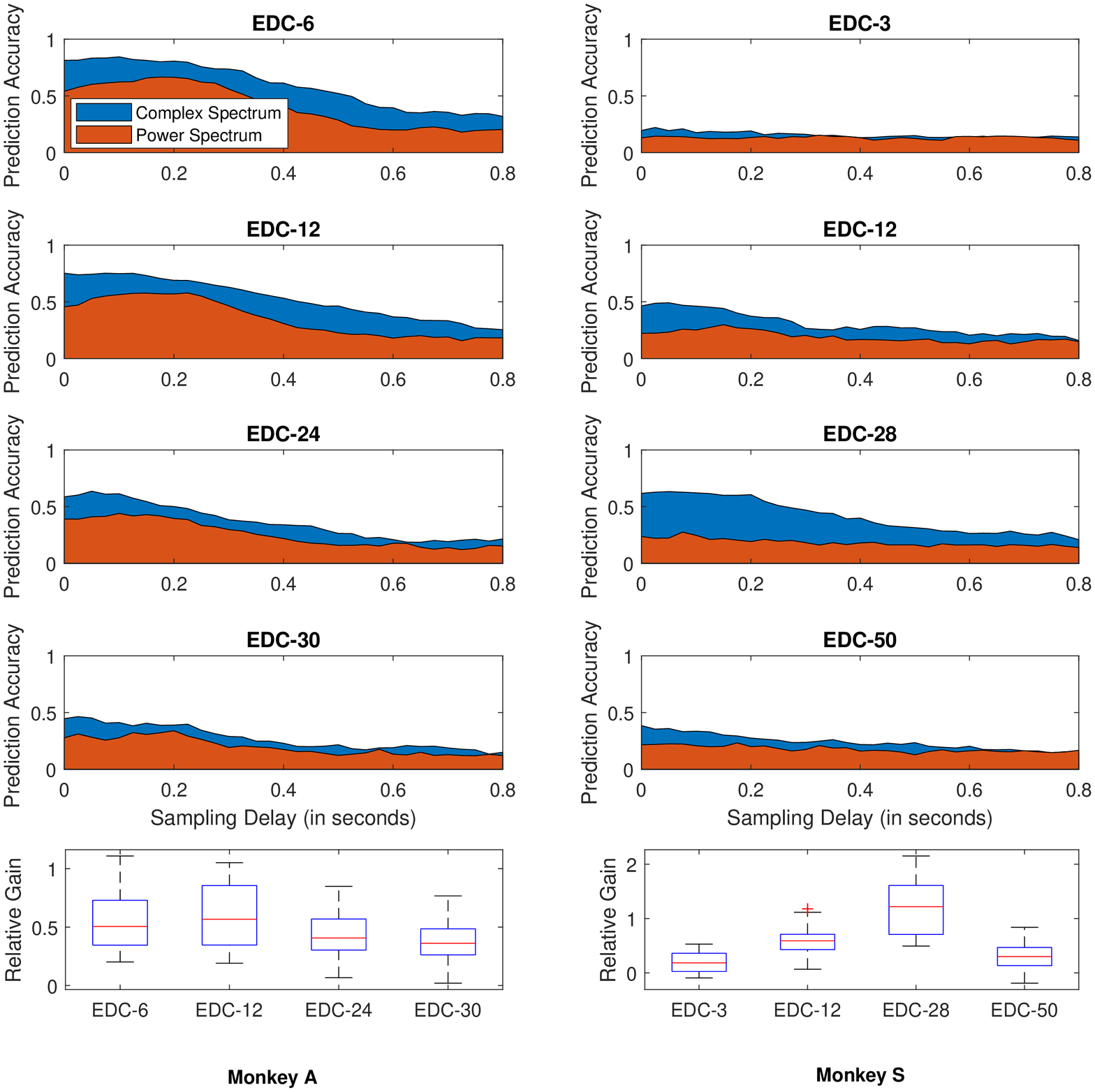}
\caption{Performance across sampling delays for representative EDCs. The sampling window size is set to $200$ milliseconds while the sampling delay is varied from $0$ to $800$ milliseconds. The remaining free parameters remain the same as in caption of Fig.~\ref{power_spectrum_depths}. The box plots depict the distribution of the relative performance gain, computed with respect to the performance achieved by the power spectrum-based decoder, across the sampling delays. 
}
\label{sampling_delay}
\end{figure}

We are next interested in the impact of the different epochs of the memory period on the dynamics of the decision making process as we try to identify which parts of the memory period carry the largest portions of information relevant for decoding eye movement directions.
Therefore, we evaluate the impact of the sampling window size and the sampling delay.
Fig.~\ref{sampling_window} compares the prediction performance of the complex spectrum decoder with the power spectrum decoder (for the same number of Fourier coefficients) at several representative EDCs in both Monkeys, for increasing sampling window sizes (expressed in milliseconds) and sampling delay of $0$; we have used the same representative EDCs to produce the analyses in Fig.~\ref{sampling_delay} 
where we show the performance of both decoders for fixed sampling window size of $200$ milliseconds and varying sampling delay, relative to the beginning of the memory period (i.e., after the fixation central target goes off, see also Fig.~\ref{experiment}).
The remaining free parameters remain the same as in prior analyses (i.e., $L=4$, $P=187$ for the complex and $P=100$ for the power spectrum decoder).
The above analyses indicate that the main part of the total information relevant for the decision making process, i.e., the direction in which the monkeys' eyes are going to saccade, and therefore, relevant for decoding, is stored in the first half of the memory period.
Specifically, Fig.~\ref{sampling_window} shows that both decoders peak halfway through the memory period, reaching saturation and remaining stable after that.
In other words, the dynamics of the system is determined early on during the memory period and this conclusion remains valid across different EDCs and across different subjects.

In Fig.~\ref{sampling_delay}  we observe a degradation of performance for both decoders as we move deeper into the memory period, clearly showing a decay of the information relevant to the decoding.
Furthermore, we see that the proposed decoder based on the complex Fourier spectrum peaks very soon after the beginning of the memory period, at around $100$ milliseconds of delay after which it gradually starts to deteriorate.
The amplitude-based decoder behaves similarly: it peaks slightly later at around $200$ milliseconds after which it starts to degrade; this also explains the steeper slope of the complex spectrum decoder before reaching saturation, compared to the slope of the power spectrum decoder, observed in Fig.~\ref{sampling_window}.
We also observe that the power degrades faster than the complex spectrum decoder; this is intuitively expected since the complex spectrum includes both the amplitude and the phase information and the addition of phase information leads to slower, more gradual degradation.


\section{Discussion}
\label{sec:conc}
We conclude the paper by summarizing its main contribution, findings and their potential practical implications in designing BMIs.

We formulated the problem of decoding movement goals from LFP signals in statistically optimal framework and proposed a novel eye movement direction decoder based on the low-pass filtered Fourier spectrum of the LFP signals \cite{Banerjee1}.
Unlike popular LFP spectrum-based decoders that rely on the power spectrum of the Fourier representation, the method proposed here naturally includes both the amplitude and the phase information by using the complex Fourier spectrum as a feature in the decoding space.
As a result, the decoder shows substantial fidelity improvement in eye movement decoding as compared with amplitude-based decoders and it peaks at $94\%$. 
The performance gain suggests that the phase information in the lowest frequencies of the LFP signal contains vital information relevant for the memory-driven decision making process of visual saccades which can be considered as an insight of fundamental importance, especially when related to previous works that have managed to identify that different frequency bands in neural recordings yield various pieces of information about various intended motor actions, despite representing different biophysiological phenomena.
In particular, this finding is in line with previous reports \cite{Pesaran:2002cfa} that have found out that higher frequency bands of the LFP power spectrum ($>50$ Hz) contain information relevant for motor decoding;
this in addition justifies the usage of the full power spectrum of the LFP as in \cite{Markowitz18412}, including the high frequency bands, when decoding the intended motor actions using power spectrum-based decoder, see also the recent review \cite{Pesaran:2018} and references therein.
Casting the problem in a minimax-optimal framework, has simply showed us that when both the amplitude and phase information from the LFP spectrum are considered, it is sufficient to use only the few lowest frequencies to generate the feature space for decoding.

An alternative to spectrum-based decoders are the time-domain decoders that use low-pass filtered version of the LFP, also known as Local Motor Potential (LMP, see \cite{Stavisky:2015fo}) as a feature for decoding intended motor actions.
Although in this paper, we have only focused on spectral features and spectrum-based decoders, the approach we propose can be also seen in the line of LMP-based time-domain decoding.
Specifically, the LMP is a low-pass filtered signal obtained by smoothing out the LFP signal using moving average filter in time-domain; the smoothing has again been applied heuristically, without deeper theoretical justification.
In similar fashion, our proposed decoder effectively performs low-pass filtering of the LFP signal, using frequency domain filter based on Pinsker's theorem which as stated earlier, is grounded in the theory of minimax-optimal estimation.

We have also investigated various other aspects of the eye movement decoding process, such as the performance of the decoder across depths and targets. 
The obtained results are consistent with previous analyses \cite{Markowitz18412} and they show 1) performance degradation with increasing electrode depth, and 2) noticeable performance disparity between ipsilateral and contralateral eye movements.
We have also studied the impact of sampling window size and sampling delay within the memory period and we have concluded that most of the information pertinent to the decision making process and, thus, to the decoding is stored within the first half of the memory period. 
This observation, combined with the fact that when combining amplitude and phase information for decoding only the low frequency band of the LFP remains relevant, can potentially have significant impact on the practical design of BMIs as it shows that the dynamics of the decision making has already been determined soon after the beginning of the memory period.
This finding suggests that the time necessary for an intended motor action to be reliably decoded can be substantially reduced---a finding with potential implications for the practical design and application of BMIs in mission-critical, tactical or public safety scenarios as no extensive memory-period sampling is necessary to determine the intended motor action with high precision.

\ifCLASSOPTIONcaptionsoff
  \newpage
\fi

\bibliographystyle{IEEEtranTCOM}
\bibliography{bare_jrnl}

\end{document}